\pdfoutput=1

\documentclass[11pt]{article}

\usepackage[final]{acl}

\usepackage{times}
\usepackage{latexsym}
\usepackage{graphics}
\usepackage{graphicx}
\usepackage{color}
\usepackage{colortbl}
\usepackage{amssymb}
\usepackage{amsmath}
\usepackage{bbm}
\usepackage{booktabs}
\usepackage{microtype}
\usepackage{multirow}
\usepackage{footmisc}
\usepackage{soul}
\usepackage{arydshln}
\usepackage{wasysym}
\usepackage{array}
\usepackage{tabularx}
\usepackage{amssymb}
\usepackage{pifont}

\usepackage{algpseudocode}
\usepackage{algorithm}
\usepackage{hyperref}
\usepackage{tikz}
\usepackage{xspace}
\usepackage{url}
\usepackage{enumitem}
\usepackage{comment}
\newtheorem{definition}{Definition}
\usepackage [english]{babel}
\usepackage [autostyle, english = american]{csquotes}
\MakeOuterQuote{"}

\usepackage{tikz}
\usepackage[edges]{forest}
\definecolor{hidden-draw}{RGB}{149,201,219}
\definecolor{tree-level-1}{RGB}{245,20,85}
\definecolor{tree-level-2}{RGB}{246,86,118}
\definecolor{tree-level-3}{RGB}{248,177,193}
\definecolor{tree-leaf}{RGB}{176,230,198}
\definecolor{lightblue}{RGB}{170, 216, 240}

\usepackage{subcaption}
\usepackage{pgfplots}
\pgfplotsset{compat=1.17}
\usetikzlibrary{pgfplots.groupplots}

\title{Large Language Models for Anomaly and Out-of-Distribution Detection:\\ A Survey}

\author{
  Ruiyao Xu\textsuperscript{\ding{171}} \and
  Kaize Ding\textsuperscript{\ding{171}} \quad  \\
  \textsuperscript{\ding{171}}Department of Statistics and Data Science, Northwestern University\\
  {ruiyaoxu2028@u.northwestern.edu}, {kaize.ding@northwestern.edu} \quad
}

\begin{document}
\maketitle
\begin{abstract}
Detecting anomalies or out-of-distribution (OOD) samples is critical for maintaining the reliability and trustworthiness of machine learning systems. Recently, Large Language Models (LLMs) have demonstrated their effectiveness not only in natural language processing but also in broader applications due to their advanced comprehension and generative capabilities. The integration of LLMs into anomaly and OOD detection marks a significant shift from the traditional paradigm in the field. This survey focuses on the problem of anomaly and OOD detection under the context of LLMs. We propose a new taxonomy to categorize existing approaches into two classes based on the role played by LLMs. Following our proposed taxonomy, we further discuss the related work under each of the categories and finally discuss potential challenges and directions for future research in this field. We also provide an up-to-date reading list \footnote{\url{https://github.com/rux001/Awesome-LLM-Anomaly-OOD-Detection}} of relevant papers. 
\end{abstract}

\section{Introduction}
Most machine learning models operate under the closed-set assumption \citep{closedset}, where the test data is assumed to be drawn i.i.d. from the same distribution as the training data. However, in real-world applications, this assumption often cannot hold, as test examples can come from distributions not represented in the training data. These instances, known as anomalies or out-of-distribution (OOD) samples, can severely degrade the performance and reliability of existing models \cite{yang2024generalized}. To build robust AI systems, methods including probabilistic approaches \citep{lee2018simple, mahalanobis18jesp} and recent deep learning techniques \citep{pang2021deep, yang2024generalized} have been explored to detect these unknown instances across various domains, such as fraud detection in finance and fault detection in industrial systems \citep{HILAL2022116429, liu2024deep}.

\begin{figure}[tbp]
    \centering
    \includegraphics[width=\columnwidth]{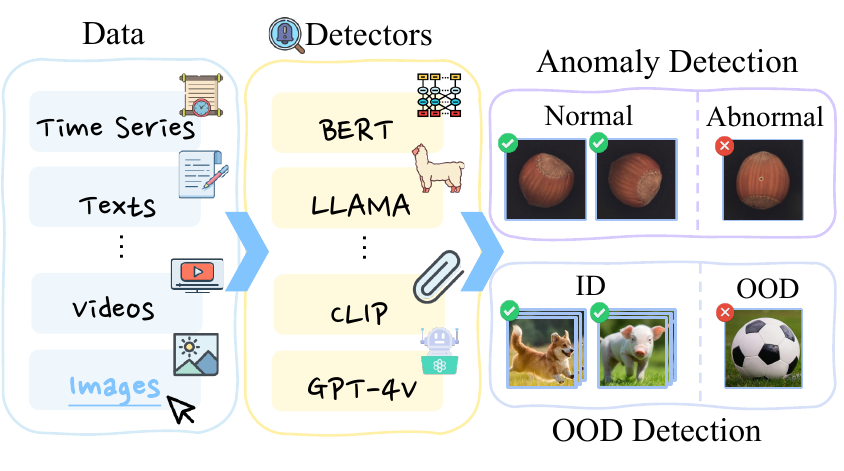}
    \caption{A simple illustration of leveraging LLMs for vision anomaly and OOD detection.}
    \label{fig:framework}
\end{figure}

Large Language Models, such as GPT-4 \cite{achiam2023gpt} and LLaMA \cite{touvron2023a}, have demonstrated remarkable capabilities in language comprehension and generation. To further harness the potential of LLMs beyond text data, there is also a growing interest in extending them to multi-modal tasks such as vision-language understanding and generation \cite{wang2024visionllm}, evolving them into Multimodal LLMs (MLLMs) \cite{yin2023survey}. Given the zero- and few-shot reasoning capabilities of LLMs and MLLMs, researchers try to apply these models to anomaly and out-of-distribution (OOD) detection, as illustrated in Figure \ref{fig:framework}, yielding promising detection results. 

Remarkably, the emergence of LLMs has fundamentally changed the learning paradigm in this field.
In the meantime, while leveraging LLMs to solve the problem of anomaly and OOD detection has drawn much attention, this field remains underexplored, highlighting the need for a comprehensive survey to analyze the emerging challenges and systematically review the rapidly expanding works. Recently, \citet{salehi2021unified} and \citet{yang2024generalized}  present unified frameworks for OOD detection but do not delve into the utilization of LLMs. While \citet{su2024large} review some small-sized language models for forecasting and anomaly detection, they neither cover the usage of LLMs with emergent abilities nor discuss OOD detection. A recent survey by \citet{miyai2024generalized} summarizes works on anomaly and OOD detection in vision using vision-language models but neglects other data modalities. Therefore, we aim to conduct a systematic survey that covers both anomaly and OOD detection across various data domains, concentrating on how LLMs are used in existing works. 

In this survey, we propose a novel taxonomy that focuses on how LLMs can profoundly impact anomaly and OOD detection in two fundamental ways, as illustrated in Figure \ref{fig:taxonomy}: \ding{182}  \textbf{LLMs for Detection (\S \ref{sec:detection}):} We provide a detailed review of existing methods that leverage LLMs as detectors for identifying anomalies and OOD instances; and \ding{183}\textbf{LLMs for Generation (\S \ref{sec:generation}):} We also review methods that utilize LLMs' emergent abilities, advanced semantic understanding, and vast knowledge to generate augmented data and explanations. At the end (\textbf{\S \ref{sec:datasets}} and \textbf{\S \ref{sec:future}}), we also summarize widely used datasets and outline future research directions, in order to provide a better understanding of anomaly and OOD detection in the era of LLMs and shed light on the following research.

\tikzstyle{my-box}=[
    rectangle,
    draw=hidden-draw,
    rounded corners,
    text opacity=1,
    minimum height=2em,
    minimum width=6em,
    inner sep=2pt,
    align=center,
    fill opacity=.5,
    font=\large,
]
\tikzstyle{leaf}=[
    my-box, 
    minimum height=2em,
    fill=lightblue!40, 
    text=black, 
    align=center,
    font=\normalsize,
    inner xsep=2pt,
    inner ysep=4pt,
]
\tikzstyle{middle}=[
    my-box, 
    draw=hidden-draw,
    minimum width=4em,
    minimum height=2em, 
    text=black, 
    align=center,
    font=\normalsize,
    inner xsep=1pt,
    inner ysep=1pt,
]

\begin{figure*}[tp]
    \centering
    \resizebox{\textwidth}{!}{
        \begin{forest}
            forked edges,
            for tree={
                grow=east,
                reversed=true,
                anchor=base west,
                parent anchor=east,
                child anchor=west,
                base=left,
                font=\normalsize,
                rectangle,
                draw=hidden-draw,
                rounded corners,
                align=center,
                minimum width=5em,
                edge+={darkgray, line width=1pt},
                s sep=3pt,
                inner xsep=2pt,
                inner ysep=3pt,
                ver/.style={rotate=90, child anchor=north, parent anchor=south, anchor=center},
            },
            where level=1{text width=4.7em,font=\normalsize,}{},
            where level=2{text width=5.5em,font=\normalsize,}{},
            where level=3{text width=5.5em,font=\normalsize,}{},
            where level=4{text width=6.5em,font=\normalsize,}{},
[
                LLMs for Anomaly and OOD Detection, ver
                [
                    LLMs for\\ Detection, middle
                    [
                        Prompting,text width=5.8em, middle
                        [
                        w/o Tuning,text width=8.3em,middle
                        [
                            SIGLLM~\cite{alnegheimish2024large}{,} LLMAD~\cite{liu2024large}{,} 
                            LogPrompt \\ ~\cite{liu2024logprompt}{,} 
                            LAVAD ~\cite{zanella2024harnessing}{,}
                            LLM-Monitor~\cite{elhafsi2023semantic}{,}\\
                            GPT-4V-AD ~\cite{zhang2023exploring}{,}
                            \cite{cao2023towards}, leaf, text width=38em
                        ]
                        ]
                        [
                        w/ Tuning,text width=8.3em,middle
                        [
                            Tabular~\cite{li2024anomaly}{,} Myriad~\cite{li2023myriad}{,} AnomalyGPT~\cite{zhang2023exploring}, leaf, text width=38em
                        ]
                        ]
                    ]
                    [
                        Contrasting,text width=5.8em,middle
                        [
                        w/o Tuning,text width=8.3em,middle
                        [
                            \cite{fort2021exploring}{,}
                            ZOC~\cite{esmaeilpour2022zero}{,} NegLabel~\cite{jiang2024negative}{,}\\
                            CLIPScope ~\cite{fu2024clipscope}{,} 
                            WinCLIP ~\cite{jeong2023winclip}{,} \\
                            AnoCLIP~\cite{deng2023anovl}{,} CLIP-AD~\cite{chen2023clip}{,} 
                            MCM ~\cite{ming2022delving}{,}\\ ~\cite{miyai2023zero}{,} 
                            SETAR~\cite{li2024setar}, leaf, text width=38em
                        ]
                        ]
                        [
                        w/ Tuning,text width=8.3em,middle
                        [
                            LoCoOp~\cite{miyai2024locoop} AnomalyCLIP~\cite{zhou2024anomalyclip}{,} 
                            InCTRL \\ ~\cite{zhu2024toward}{,} MVFA ~\cite{huang2024adapting}{,} 
                            ID-like~\cite{bai2024id}{,} \\
                            NegPrompt~\cite{li2024learning}{,} 
                            CLIPN~\cite{wang2023clipn}{,} \\
                            LSN~\cite{nie2024outofdistribution}{,} 
                            MCM-PEFT~\cite{ming2024does}, leaf, text width=38em
                        ]
                        ]
                    ]
                ]
                [
                    LLMs for\\ Generation,middle
                    [
                    Augmentation,middle,text width=5.8em
                    [
                        Text Embedding,text width = 8.2em,middle
                        [
                            LogGPT~\cite{qi2023loggpt}{,} LogFit~\cite{LogFiT}{,} \\
                            ~\cite{liu-etal-2024-good}{,} ~\cite{zhang2024your}, leaf, text width=38.2em
                        ]
                    ]
                    [
                        Pseudo Label,text width = 8.2em,middle
                        [
                            EOE~\cite{cao2024envisioning}{,} PCC~\cite{huang2024out}{,}\\
                            TOE~\cite{park2023on}{,} CoNAL~\cite{xu-etal-2023-contrastive},
                            leaf, text width=38.2em
                        ]
                    ]
                    [
                        Textual Description,text width = 8.2em,middle
                        [
                            TagFog ~\cite{chen2024tagfog}{,} ALFA~\cite{zhu2024llms}{,} 
                            ~\cite{dai2023exploring}, leaf, text width=38.2em
                        ]
                    ]
                    ]
                    [
                    Explanation,middle,text width=5.8em
                    [
                        Holmes-VAD~\cite{zhang2024holmes}{,} AnomalyRuler~\cite{yang2024follow}{,} 
                        \\ VAD-LLaMA ~\cite{lv2024video}{,}
                        AESOP~\cite{sinha2024real}, leaf, text width=47.78em
                    ]
                ]
            ]
]
        \end{forest}
    }
    \caption{Taxonomy of methods utilizing LLMs for anomaly and OOD detection tasks.}
    \label{fig:taxonomy}
\end{figure*}

\section{Preliminaries}
\label{sec:preliminaries}
\noindent\textbf{Large Language Models.}
Large language models (LLMs) generally refer to Transformer-based pre-trained language models with hundreds of billions of parameters or more. Early LLMs like BERT \cite{devlin2018bert} and RoBERTa \cite{liu2019roberta} utilize an encoder-only architecture, excelling in text representation learning \cite{bengio2013representation}. Recently, the focus has shifted toward models aimed at natural language generation, often using the “next token prediction” objective as their core task. Examples include T5 \cite{raffel2020exploring} and BART \cite{lewis2019bart}, which employ an encoder-decoder structure, as well as GPT-3 \cite{brown2020language}, PaLM \cite{chowdhery2023palm}, and LLaMA \cite{touvron2023a}, which are based on decoder-only architectures. Advancements in these architectures and training methods have led to superior reasoning and emergent abilities, such as in-context learning\cite{brown2020language} and chain-of-thought reasoning \cite{wei2022chain}.

\noindent\textbf{Multimodal Large Language Models.}
The remarkable abilities of Large Language Models (LLMs) have inspired efforts to integrate language with other modalities, with a particular focus on combining language and vision. Notable examples of Multimodal Large Language Models (MLLMs) include CLIP \cite{radford2021learning}, BLIP2 \cite{li2023blip}, and Flamingo \cite{alayrac2022flamingo}, which were pre-trained on large-scale cross-modal datasets comprising images and text. Models like GPT-4(V) \cite{openai2023} and Gemini \cite{team2023gemini} showcase the emergent abilities of Multimodal LLMs, significantly improving the performance of vision-related tasks. 

\subsection{Problem Definition}

With LLMs advancing in zero-shot and few-shot learning, the general pipeline of anomaly and out-of-distribution  (OOD) detection methods shifts to adapt pre-trained LLMs for detection without extensive training. This shift challenges traditional definitions of anomaly and OOD detection, as the conventional train-test paradigm may not always apply. Following previous studies~\cite{miyai2024generalized, yang2024generalized}, we propose to redefine anomaly and OOD detection under the context of LLMs and highlight the differences between the two problems as follows:

\begin{definition}\textbf{LLM-based Anomaly Detection:}
Given a test dataset $\mathcal{D}_{test}=\{x_1, \cdots, x_{n}\}$, where each sample $x_i$ is drawn from distribution~$\mathbb{P}^{in}$ or $\mathbb{P}^{out}$. The objective of LLM-based Anomaly Detection is to use a pre-trained LLM as the backbone and develop a detection model $f_{LLM}(\cdot)$ to predict whether each sample $x' \in \mathcal{D}_{test}$ belongs to $\mathbb{P}^{out}$, where $\mathbb{P}^{out}$ has covariate shift with $\mathbb{P}^{in}$
\end{definition}

\begin{definition}\textbf{LLM-based OOD Detection:}
Given a test dataset $\mathcal{D}_{\text{test}}=\{x_1, \cdots, x_{n}\}$, where each sample $x_i$ is drawn from distribution $\mathbb{P}^{\text{in}}$ or $\mathbb{P}^{\text{out}}$, and a known ID class set $\mathcal{C}=\{c_1, \cdots, c_{k}\}$. The objective of LLM-based OOD Detection is to use a pre-trained LLM as backbone and develop detection model $f_{\text{LLM}}(\cdot)$ to predict whether each sample $x' \in \mathcal{D}_{\text{test}}$ belongs to $\mathbb{P}^{\text{out}}$, where $\mathbb{P}^{\text{out}}$ has semantic shift with $\mathbb{P}^{\text{in}}$. If not, $x'$ will be classified into one of the classes in $\mathcal{C}$.
\end{definition}

\noindent\textbf{Discussions.}
The distinction between anomaly detection and OOD detection in the context of LLMs highlights the unique challenges posed by covariate and semantic shifts. \textit{Anomaly detection} aims to identify subtle deviations within the data that may not involve a complete change in the underlying class or concept, such as detecting defects or irregularities in industrial processes. In contrast, \textit{OOD detection} focuses on identifying instances that do not belong to any of the known ID classes at the object level, such as recognizing a dog when the only provided ID class is cat. This differentiation underscores the need for tailored approaches for each detection task.

\section{LLMs for Detection}
\label{sec:detection}
\begin{figure}[tbp]
    \centering
    \includegraphics[width=\columnwidth]{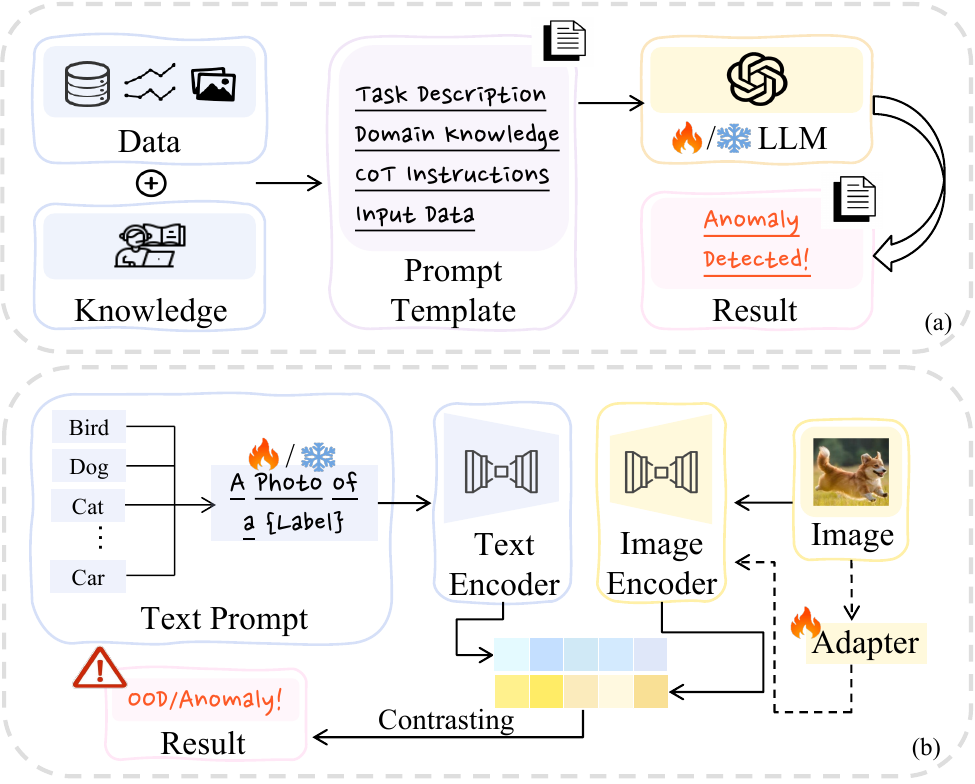}
    \caption{The illustration of two approaches in (\S \ref{sec:detection}): (a) Prompting-based Detection and (b) Contrasting-based Detection.}
    \label{fig:detection}
\end{figure}

The primary objective of this section is to explore existing works that utilize LLMs' inherent knowledge to detect anomalies or OOD samples. Under this line of research, approaches can be categorized into two classes as illustrated in Figure \ref{fig:detection}: \ding{182} Prompting-based Detection methods, which involve directly prompting LLMs to generate language responses that include detection results; \ding{183} Contrasting-based Detection methods, which focus on multimodal scenarios, using MLLMs pre-trained with a contrastive objective as detectors.

\subsection{Prompting-based Detection}
The general pipeline for prompting-based detection methods consists of two primary stages: (i) constructing a structured prompt template with instruction prompt \(\mathcal{P}\) and input data \(\mathcal{X}\); and (ii) feeding the template-based prompt $\hat{X}$ into LLMs to generate a language response. The function $\texttt{Parse}(\cdot)$ is then applied to extract the detection results. Depending on the scenario, the LLM can either be frozen or fine-tuned, denoted as \( f_{\textsc{LLM}}^{\heartsuit} \) or \( f_{\textsc{LLM}}^{\clubsuit} \), respectively. This process can be summarized as follows:
\begin{equation*}
\begin{split}
    \text{Prompt Construction:}&\quad \hat{X} = \texttt{Template}(\mathcal{X}, \mathcal{P}), \\
    \text{Detection:}&\quad \tilde{Y} = \texttt{Parse}\left(f_{\textsc{LLM}}^{\heartsuit/\clubsuit}(\hat{X})\right)
\end{split}
\end{equation*}
Note that the prompting-based approach primarily addresses the anomaly detection task. OOD detection research has not yet widely adopted prompting to directly identify OOD samples.

\subsubsection{Detection without LLM Tuning}
Since some approaches do not require additional tuning, they mainly focus on employing various prompt engineering techniques \cite{sahoo2024systematic} to guide LLMs to produce better detection results. To design suitable prompts for anomaly detection, researchers have employed a combination of various prompt techniques, such as \textit{role-play prompting}~\cite{wu2023large}, \textit{in-context learning}~\cite{brown2020language}, and \textit{chain-of-thought (CoT) reasoning} \cite{wei2022chain}, to create effective prompt templates. Studies such as SIGLLM \cite{alnegheimish2024large}, LLMAD \cite{liu2024large}, and LogPrompt \cite{liu2024logprompt} focus on time series and log data. SIGLLM \cite{alnegheimish2024large} investigates two distinct pipelines for using LLMs in time series anomaly detection: one directly prompts an LLM with specific role-play instructions to identify anomalous elements in given data, and the other uses the LLM’s forecasting ability to detect anomalies by comparing original and forecasted signals, where discrepancies indicate anomalies. LLMAD \cite{liu2024large} incorporates in-context learning examples retrieved from a constructed database and CoT prompts that inject domain knowledge of time series. LogPrompt \cite{liu2024logprompt} explores three prompting strategies for log data: self-prompt, CoT prompt, and in-context prompt, demonstrating that the prompt with CoT techniques outperforms other prompting strategies. The tailored CoT prompt for log data includes a specific task instruction, i.e. "classify the given log entries into normal and abnormal categories", and step-by-step rules for considering given data as anomalies.   

Unlike time series and log data which can be directly converted into raw text data, other data modalities, such as videos and images, require additional processing to be transformed into a format that LLMs can understand. For instance, LAVAD \cite{zanella2024harnessing} first exploits a captioning model to generate a textual description for each video frame and further uses an LLM to summarize captions within a temporal window. This summary is then used to prompt the LLM to provide an anomaly score for each frame. LLM-Monitor \cite{elhafsi2023semantic} uses an object detector to identify objects in video clips and then designs specific prompt templates incorporating CoT and in-context examples to query LLMs for anomaly detection.

With the integration of multimodal understanding into LLMs, these models are now capable of comprehending various modalities beyond text, enabling more direct applications for anomaly detection across a wide range of data types. \citet{cao2023towards} conduct comprehensive experiments and analyses using GPT-4V(ision) for anomaly detection across various modality datasets and tasks. To enhance GPT-4V’s performance, they also incorporate different types of additional cues such as class information, human expertise, and reference images as prompts. Similarly, GPT-4V-AD \cite{zhang2023exploring} employs GPT-4V as the backbone, designing a general prompt description for all image categories and injecting specific image category information, resulting in a specific output format for each region with respective anomaly scores.

\subsubsection{Detection with LLM Tuning}
Directly prompting frozen LLMs for anomaly or OOD detection results across various data types often yields suboptimal performance due to the inherent modality gap between text and other data modalities.  As a result, additional training and fine-tuning on LLMs for downstream detection tasks has become a prevalent research trend. Unfortunately, fine-tuning entire LLMs is often computationally expensive and poses significant challenges. Therefore, parameter-efficient fine-tuning (PEFT) has been extensively employed instead. For example, Tabular \cite{li2024anomaly} designs a prompt template to query the LLM to output anomalies based on given converted tabular data. To better adapt the LLM for anomaly detection at the batch level, they apply Low-Rank Adaptation (LoRA), using a synthetic dataset with ground truth labels in a supervised manner.

To enhance LLMs for localization understanding and adapting to industrial tasks, AnomalyGPT \cite{zhang2023exploring} first derives localization features from a frozen image encoder and image decoder and these features are then fed to a tunable prompt learner. Without fine-tuning the entire LLM, they fine-tune the prompt learner with LoRA to significantly reduce computational costs. Myriad \cite{li2023myriad} employs Mini-GPT-4 as the backbone and integrates a trainable encoder, referred to as Vision Expert Tokenizer, to embed the vision expert’s segmentation output into tokens that the LLM can understand. With expert-driven visual-language extraction, Myriad can generate accurate anomaly detection descriptions. 

\subsection{Contrasting-based Detection} 
In this section, we focus on MLLMs, such as CLIP, which are pre-trained with an image-text contrastive objective and learn by pulling the paired images and texts close and pushing others far away in the embedding space. The zero-shot classification ability of these models further builds the foundation for contrasting-based anomaly and OOD detection methods: (i) given an image \( x_i \) and a text prompt \( f \) with a target class set \( C \), CLIP extracts image features \( h \in \mathbb{R}^D \) using an image encoder \( f_{\text{img}} \), and text features \( e_j \in \mathbb{R}^D \) using a text encoder \( f_{\text{text}} \) with a prompt template for each class \( c_j \in C \), and (ii) the similarity between \( h \) and each \( e_j \) is usually used as an important component in the score function \(f_{\text{score}}\) for deciding whether \( x_i \) is an anomaly or OOD sample. 
This process can be summarized as follows:
\begin{equation*}
\begin{split}
    \text{Feature Extraction:}&\quad h = f_{\text{img}}(x_i), \\and 
    &\quad e_j = f_{\text{text}}(\text{prompt}(c_j)), \\
    \text{Detection:}&\quad \tilde{Y} = f_{\text{score}}\left(\cos(h, e_j)\right)
\end{split}
\end{equation*}
We further categorize contrasting-based detection methods into two classes depending on whether there exists additional training and fine-tuning.

\subsubsection{Detection without LLM Tuning} 
Anomaly and OOD detection problems can indeed be understood as classification problems. Therefore, pretrained MLLMs like CLIP, with strong zero-shot classification ability, can serve as detectors themselves. By using only ID or normal prompts, CLIP can be leveraged for both OOD and anomaly detection tasks. Despite the promise, existing CLIP-like models perform zero-shot classification in a closed-world setting. That is, it will match an input into a fixed set of categories, even if it is irrelevant \cite{ming2022delving}. To address this, one approach involves designing effective post-hoc score functions that rely solely on ID or normal class labels. Alternatively, some researchers incorporate anomaly or OOD class information into the text prompts, allowing the model to match OOD or abnormal images to paired prompts. 

\begin{itemize}[leftmargin=*,itemsep=0.2pt,topsep=1.pt]
    \item \textit{Without Anomaly/OOD Prompts.} 
For \textbf{anomaly detection}, WinCLIP \cite{jeong2023winclip} initially investigates a one-class design by using only the normal prompt “normal [o]” where [o] represents object-level label, i.e "bottle", and defining an anomaly score as the similarity between vectors derived from the image encoder and normal prompts. However, this one-class design yields poorer results compared to a simple binary zero-shot framework, CLIP-AC \cite{jeong2023winclip}, which adapts CLIP with two class prompts: “normal [o]” vs. “anomalous [o]”. These results highlight that the one-class design is less effective for anomaly detection, and as a result, anomaly detection research generally does not follow this line with only normal prompts.

For \textbf{OOD detection}, to address the challenges of using only in-distribution (ID) class information while avoiding the matching of OOD inputs to irrelevant ID classes, one notable approach is the Maximum Concept Matching (MCM) framework proposed by \cite{ming2022delving}. This method is not limited to CLIP and can be generally applicable to other pre-trained models that promote multi-modal feature alignment. They view the textual embeddings of ID classes as a collection of concept prototypes and define the maximum concept matching (MCM) score based on the cosine similarity between the image feature and the textual feature. Following the idea of MCM, several subsequent works focus on improving OOD detection results by either adding a local MCM score or modifying weights in the original MCM framework, such as \cite{miyai2023zero} and \cite{li2024setar}.

    \item \textit{With Anomaly/OOD Prompts.} 
\citet{fort2021exploring} first investigate using CLIP for \textbf{OOD detection} and demonstrate encouraging performance. However, in their setup, they include the candidate labels related to the actual OOD classes and utilize this knowledge as a very weak form of outlier exposure, which contradicts the open-world assumption. Therefore, after this work, researchers aim to leverage pseudo-OOD labels in the text prompt instead of using actual OOD labels. The earliest work under this idea is ZOC \cite{esmaeilpour2022zero} which trains a text description generator on top of CLIP’s image encoder to dynamically generate candidate unseen labels for each test image. The similarity of the test image with seen and generated unseen labels is used as the OOD score. Instead of training an additional text decoder, NegLabel \cite{jiang2024negative} and CLIPScope \cite{fu2024clipscope} rely on auxiliary datasets to gather potential OOD labels. CLIPScope gathers nouns from open-world sources as potential OOD labels and uses them in designed prompts to ensure maximal coverage of potential OOD samples. NegLabel employs the NegMining algorithm to select high-quality negative labels with sufficient semantic differences from ID labels. Recent work utilizes the emergent abilities of LLMs to generate reliable OOD labels, such as \cite{cao2024envisioning}, \cite{huang2024out}, \cite{park2023on}, and \cite{xu-etal-2023-contrastive}. 

For contrasting-based \textbf{anomaly detection}, following the simple binary zero-shot framework, CLIP-AC \cite{jeong2023winclip}, which adapts CLIP with two class prompts: “normal [o]” vs. “anomalous [o]”, many subsequent research emerge. While using the default prompt has demonstrated promising performance, similar to the prompt engineering discussion around GPT-3 \cite{brown2020language}, researchers have observed that performance can be significantly improved by customizing the prompt text. Models like WinCLIP \cite{jeong2023winclip} and AnoCLIP \cite{deng2023anovl} use a Prompt Ensemble technique to generate all combinations of pre-defined lists of state words per label and text templates. After generating all combinations of states and templates, they compute the average of text embeddings per label to represent the normal and anomalous classes. In practice, more descriptions in prompts do not always yield better performance. Therefore, CLIP-AD \cite{chen2023clip} proposes Representative Vector Selection (RVS), from a distributional perspective for the
design of the text prompt, broadening research opportunities beyond merely crafting adjectives. 
\end{itemize}

\subsubsection{Detection with LLM Tuning}
Following the similar detection pipeline of methods without LLM tuning, researchers propose to employ prompt tuning or adapter tuning techniques to eliminate the need for manually crafting prompts and enhance the understanding of local features of images. Additionally, by incorporating a few ID or normal images during training or inference phases, some methods transition into few-shot scenarios.

\begin{itemize}[leftmargin=*,itemsep=0.2pt,topsep=1.pt]
    \item \textit{LLM Adapter-Tuning.}
Adapter-tuning methods involve integrating additional components or layers into the model architecture to facilitate better alignment or localization \cite{hu2023llm}. This approach is significantly useful for \textbf{anomaly detection} task, because CLIP was originally designed for classifying the semantics of objects in the scene, which does not align well with the sensory anomaly detection task where both normal and abnormal samples are often from the same class of object. To reconcile this, InCTRL \cite{zhu2024toward} includes a tunable adapter layer to further adapt the image representations for anomaly detection. To better adapt to medical image anomaly detection, MVFA \cite{huang2024adapting} proposes a multi-level visual feature adaptation architecture to align CLIP’s features with the requirements of anomaly detection in medical contexts. This is achieved by integrating multiple residual adapters into the pre-trained visual encoder, guided by multi-level, pixel-wise visual-language feature alignment loss functions.

\item \textit{LLM Prompt-Tuning.} 
Manually crafting suitable prompts always requires extensive human effort. Therefore, researchers employ the idea of prompt tuning, such as CoOp \cite{zhou2022learning}, to learn a soft or differentiable context vector to replace the fixed text prompt. For \textbf{OOD detection}, most approaches rely on using auxiliary prompts to represent potential OOD textual information, and one crucial problem is to identify hard OOD data that is similar to ID samples. To solve this, \citet{bai2024id} first constructs outliers highly correlated with ID data and introduces a novel prompt learning framework for learning specific prompts for the most challenging OOD samples, which behave like ID classes. Additionally, LSN \cite{nie2024outofdistribution}, NegPrompt \cite{li2024learning}, and CLIPN \cite{wang2023clipn} all work on learning extra negative prompts to fully leverage the capabilities of CLIP for OOD detection. Unlike the other two approaches, CLIPN requires training an additional "no" text encoder using a large external dataset to get negative prompts for all classes. This auxiliary training is computationally expensive, limiting its application to generalized tasks. Also, LSN demonstrates that naive "no" logic prompts cannot fully leverage negative features. Therefore, both LSN and NegPrompt focus on training on more detailed negative prompts, while LSN also aims to develop class-specific positive and negative prompts, enabling more accurate detection. 

Instead of focusing on leveraging OOD information into the text encoder, some methods aim to perform prompt tuning to optimize word embeddings for ID labels and then use the MCM score as the detection criterion. MCM-PEFT~\cite{ming2024does} demonstrates that simply applying prompt tuning for CLIP on few-shot ID datasets can significantly improve detection accuracy. However, a primary limitation of this approach is its exclusive reliance on the features of ID classes, leading to incorrect detection when input images share a high visual similarity with the class in the prompt. To address this, LoCoOp \cite{miyai2024locoop} treats such ID-irrelevant nuisances as OOD and learns to push them away from the ID class text embeddings, preventing the model from producing high ID confidence scores for the OOD features. Additionally, \citet{lafon2024gallop} enhances detection capabilities by learning a diverse set of prompts utilizing both global and local visual representations. For \textbf{anomaly detection} which emphasizes more on learning local features, AnomalyCLIP \cite{zhou2024anomalyclip} aims to learn object-agnostic text prompts that capture generic normality and abnormality in images, allowing the model to focus on abnormal regions rather than object semantics. 

\end{itemize}
\section{LLMs for Generation}
\label{sec:generation}
In this section, we review methods that leverage LLMs as generative tools for enhancing anomaly and OOD detection. LLMs use their extensive pre-trained knowledge to generate augmented data, such as embeddings, pseudo labels, and textual descriptions, improving detection performance~\cite{li-etal-2024-empowering,ding-etal-2024-data}. Furthermore, due to their ability to understand and generate human-like text, LLMs have been explored for providing insightful explanations and analyses of detection results, aiding in interpretation, planning, and decision-making. These methods are classified into two main approaches: \ding{182} Augmentation-centric Generation and \ding{183} Explanation-centric Generation.

\subsection{Augmentation-centric Generation}

LLMs serve as effective tools for data augmentation in anomaly and OOD detection tasks by generating textual embeddings, pseudo labels, and descriptive text. The extensive pre-training of LLMs on large datasets and the autoregressive training objective endow them with superior generative capabilities. These capabilities allow LLMs to produce high-quality embeddings, create synthetic labels, and provide additional descriptive information, ultimately boosting the performance and robustness of detection models.

\subsubsection{Text Embedding-based Augmentation}
LLMs are highly effective feature extractors, producing meaningful embeddings that can be used in detection tasks. This augmentation enables detection models to capture more subtle patterns and distinctions, leading to more accurate and robust detection performance. For instance, \citet{hadadi2024anomaly} and \citet{qi2023loggpt} fine-tune pre-trained GPT models on log data and use the extracted semantic embeddings as key components for future anomaly detection.

For OOD detection in text data, a common approach involves using encoder-only LLMs to generate sentence representations that are used to compute OOD confidence scores \cite{liu-etal-2024-good}. These models are typically fine-tuned on ID data, and OOD detectors are applied to the generated representations. Recently, there has been a shift toward leveraging larger language models with decoder architectures, which provide enhanced capabilities in refining textual representations. \citet{liu-etal-2024-good} explore the use of decoder-only LLMs, such as LLaMa, incorporating fine-tuning techniques like LoRA to reduce the additional parameters. Their findings demonstrate that fine-tuned LLMs, combined with customized OOD scoring functions, significantly improve OOD detection performance. A key advantage of decoder-based LLMs is their autoregressive ability, which allows for more effective handling of sequential data. Building on this, \citet{zhang2024your} propose using the likelihood ratio between a pre-trained and fine-tuned LLM as a criterion for OOD detection, leveraging the deep contextual understanding embedded within LLMs for text data.

\subsubsection{Pseudo Label-based Augmentation}
The emergent capabilities of LLMs provide a promising approach for generating high-quality synthetic datasets, including pseudo labels for OOD samples. A significant challenge in OOD detection is the lack of labeled OOD data, which can limit model performance. Traditionally, obtaining OOD labels requires extensive human effort, but LLMs can mitigate this by generating pseudo-OOD labels through carefully designed prompts. 

For example, EOE \cite{cao2024envisioning} and PCC \cite{huang2024out} prompt LLMs to generate visually similar OOD class labels, which are then used to define a new scoring function. This approach significantly outperforms methods relying solely on known ID labels. TOE \cite{park2023on} further evaluates the generation of pseudo-OOD labels at three verbosity levels—word-level, description-level, and caption-level—using BERT, GPT-3, and BLIP-2, respectively. Results indicate that caption-level pseudo-OOD labels generated by BLIP-2, which incorporates both semantic and visual understanding, perform the best. In text data, CoNAL \cite{xu-etal-2023-contrastive} prompts LLMs to extend closed-set labels with novel examples and generates a comprehensive set of probable OOD samples. By applying contrastive confidence loss during training, the model achieves high accuracy on the ID training set while maintaining lower confidence on the generated OOD examples.

\subsubsection{\hspace{-0.1cm}Textual Description-based Augmentation}
In addition to generating pseudo labels, LLMs are also used to generate textual descriptions of both known ID classes and potential OOD samples. For example, TagFog \cite{chen2024tagfog} employs a Jigsaw strategy to generate fake OOD samples and prompts ChatGPT to create detailed descriptions for each ID class, guiding the training of the image encoder in CLIP for OOD detection. In anomaly detection tasks, it is essential for LLMs to recognize the close correlation between normal images and their respective prompts, while identifying a more distant association with abnormal prompts. This requires detailed and nuanced descriptions of normal and anomalous stages of objects. ALFA \cite{zhu2024llms} formulates prompts for LLMs to describe both normal and abnormal features for each class, and these descriptions are then used to improve the detection of abnormal objects. To avoid LLM hallucination, \citet{dai2023exploring} introduce a consistency-based uncertainty calibration method, where LLMs describe visual features for distinguishing categories in images, and the confidence score of each generation is estimated accordingly.

\begin{figure}[tbp]
    \centering
    \includegraphics[width=\columnwidth]{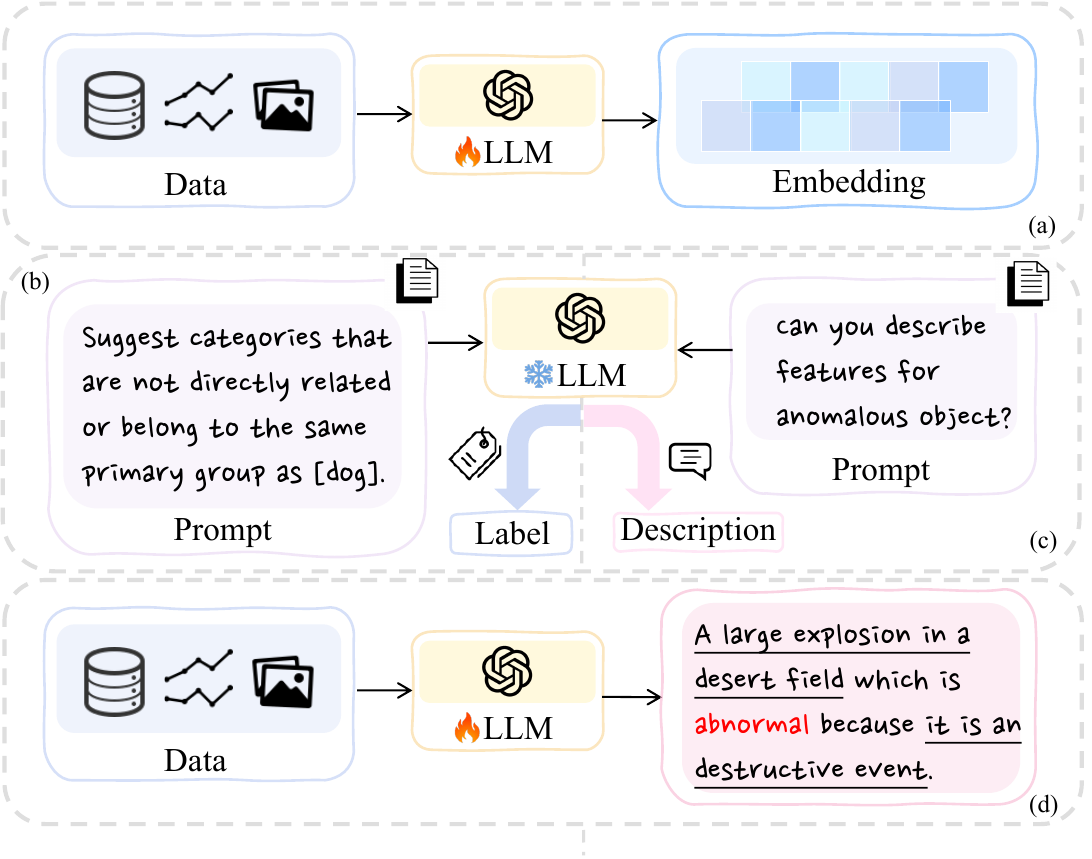}
    \caption{The illustration of four approaches in (\S \ref{sec:generation}): (a) Text Embedding-based Augmentation; (b) Pseudo Label-based Augmentation; (c) Textual Description-based Augmentation; and (d) Explanation-centric Generation.}
    \label{fig:generation}
\end{figure}

\subsection{Explanation-centric Generation}
\label{sec:explanation}
Beyond augmentation, LLMs' powerful reasoning and natural language generation abilities allow them to provide insightful explanations for anomaly and OOD detection outcomes. These explanations are especially important in safety-critical domains, such as autonomous driving, where transparency and interpretability are crucial.

For example, Holmes-VAD \cite{zhang2024holmes} trains a lightweight temporal sampler to select frames with high anomaly scores and uses an LLM to generate detailed explanations, providing clear insights into the detected anomalies. VAD-LLaMA \cite{lv2024video} generates instruction-tuning data to train the projection layer of Video-LLaMA, enabling more comprehensive explanations of anomalies. AnomalyRuler \cite{yang2024follow} employs a rule-based reasoning strategy with few-normal-shot prompting, providing interpretable, rule-driven explanations that can quickly adapt to various video anomaly detection scenarios.

Additionally, LLMs are being used in autonomous agents to guide decision-making after anomaly or OOD detection. For instance, AESOP \cite{sinha2024real} leverages the autoregressive capabilities of an LLM to provide zero-shot assessments on whether interventions are required in robotic systems after an anomaly is detected. By utilizing LLMs' generative reasoning, these systems can plan and respond to anomalies in an efficient and informed manner.

\section{Evaluation Datasets}
\label{sec:datasets}
In this section, we introduce commonly used datasets across multiple modalities in Table \ref{tab:datasets}, including images, videos, text, and time series, that serve as benchmarks for anomaly and OOD detection We also introduce widely used evaluation metrics in Appendix \ref{sec:metrics}.

\begin{table}[ht]
\centering
\small
\resizebox{\linewidth}{!}{
\begin{tabularx}{\columnwidth}{l l X}
\toprule
 & \textbf{Task Type} & \textbf{Datasets} \\
\midrule
\multirow{11}{*}{\rotatebox{90}{Images}} 
 & Anomaly & MVTec-AD~\cite{8954181} \\
 & Anomaly & VisA~\cite{zou2022spot} \\
 & Anomaly & Chest X-ray~\cite{kermany2018identifying} \\
 & Anomaly & Head CT~\cite{Kitamura2018headct} \\
 & Anomaly & LOCO~\cite{bergmann2022beyond} \\
 & OOD & ImageNet~\cite{5206848} \\
 & OOD & iNaturalist~\cite{van2018inaturalist} \\
 & OOD & Places~\cite{zhou2017places} \\
 & OOD & SUN~\cite{xiao2010sun} \\
 & OOD & Texture~\cite{texture} \\
\midrule
\multirow{3}{*}{\rotatebox{90}{Videos}} 
 & Anomaly & UCF-Crime~\cite{sultani2018real} \\
 & Anomaly & XD-Violence~\cite{wu2020not} \\
 & Anomaly & VAD-Intruct50k~\cite{zhang2024holmes} \\
\midrule
\multirow{6}{*}{\rotatebox{90}{Texts}} 
 & Anomaly & NLP-AD~\cite{li2024nlp} \\ 
 & Anomaly & AD-NLP~\cite{bejan-etal-2023-ad} \\
 & OOD & 20NG~\cite{LANG1995331} \\
 & OOD & SST-2~\cite{socher-etal-2013-recursive} \\
 & OOD & CLINC150~\cite{clinc150_570} \\
\midrule
\multirow{5}{*}{\rotatebox{90}{Times Series}} 
 & Anomaly & ODDS~\cite{Rayana:2016} \\
 & Anomaly & Yahoo~\cite{10.14778/3529337.3529354} \\
 & Anomaly & ECG~\cite{10.14778/3529337.3529354} \\
 & Anomaly & SVDB~\cite{10.14778/3529337.3529354} \\
 & Anomaly & IOPS~\cite{10.14778/3529337.3529354} \\
\bottomrule
\end{tabularx}
}
\caption{A summary of datasets for Anomaly and Out-of-Distribution (OOD) Detection by Modality.}
\label{tab:datasets}
\end{table}

    \noindent\textbf{Images.} Industrial anomaly detection is commonly evaluated using MVTec-AD~\cite{8954181} and VisA~\cite{zou2022spot}, which provide benchmarks for defect detection and localization. In the medical domain, datasets such as Chest X-ray~\cite{kermany2018identifying} and Head CT~\cite{Kitamura2018headct} are used for detecting abnormalities in clinical imaging. Logical anomaly detection is assessed with MVTec LOCO~\cite{bergmann2022beyond}. For OOD detection, which focuses on semantic shifts, large-scale datasets such as ImageNet~\cite{5206848}, iNaturalist~\cite{van2018inaturalist}, Places~\cite{zhou2017places}, and SUN~\cite{xiao2010sun} are widely used.
    
    \noindent\textbf{Videos.} UCF-Crime~\cite{sultani2018real} and XD-Violence~\cite{wu2020not} are widely used for detecting anomalous activities in surveillance footage. The VAD-Instruct50k dataset~\cite{zhang2024holmes} introduces a large-scale multimodal benchmark for video anomaly detection.
    
    \noindent\textbf{Texts.} Textual anomaly and OOD detection have gained attention with the rise of LLMs. NLP-AD~\cite{li2024nlp} and AD-NLP~\cite{bejan-etal-2023-ad} provide benchmarks for detecting anomalous patterns in textual data. Recent work, AD-LLM~\cite{yang2024ad}, evaluates LLM performance on text anomaly detection, demonstrating their superior effectiveness in this task. Textual OOD detection is commonly assessed using CLINC150~\cite{clinc150_570}, which partitions intent categories into ID and OOD classes. Additional benchmarks, such as 20 Newsgroups (20NG)~\cite{LANG1995331} and SST-2~\cite{socher-etal-2013-recursive}, are used for OOD detection in topic classification and sentiment analysis.
    
    \noindent\textbf{Time Series.} The ODDS dataset~\cite{Rayana:2016} serves as a standard benchmark for time series anomaly deteciton, while Yahoo, ECG, SVDB, and IOPS datasets~\cite{10.14778/3529337.3529354} focus on detecting anomalies in financial transactions, physiological signals, and system performance.

\section{Challenges and Future Directions}
\label{sec:future}
In this section, we briefly summarize challenges and future directions within the anomaly and OOD detection research field in the era of LLMs.

\noindent\textbf{Explainability and Trustworthiness.}
In addition to accurately detecting anomalies or OOD samples, there is an increasing trend to utilize LLMs to provide reasonable explanations and serve as agents to plan future actions. Future research should focus on developing methods to enhance the explainability of LLMs for anomaly or OOD detection, increasing the trustworthiness of LLM-based systems and facilitating their adoption in critical domains such as healthcare and finance~\citep{holzinger2019causability, guidotti2019survey}.


\noindent\textbf{Unsolvable Problem Detection.}
\citet{miyai2024unsolvable} propose Unsolvable Problem Detection (UPD), which evaluates the LLMs’ ability to recognize and abstain from answering unexpected or unsolvable input questions, aiding in preventing incorrect or misleading outputs in critical applications where the consequences of errors can be significant. Future work should focus on incorporating the concepts of OOD detection techniques for solving the UPD problems. 

\noindent\textbf{Handling Multimodal Data.}
The emergence of MLLMs capable of processing and understanding multiple data modalities offers significant potential in the field of anomaly and OOD detection~\citep{alayrac2022flamingo, li2023blip}. Future research should explore methods to better adapt LLMs to comprehend and integrate various multimodal data, thereby enhancing their ability to detect anomalies and OOD instances across diverse datasets.
\section{Conclusion}
In this survey, we examine the use of Large Language Models (LLMs) and multimodal LLMs (MLLMs) in anomaly and out-of-distribution (OOD) detection. We introduce a novel taxonomy categorizing methods into two approaches based on the role of LLMs in the architectures: detection, and generation. This taxonomy clarifies how LLMs can augment data, detect anomalies or OOD, and build explainable systems. We also summarize commonly used datasets and discuss future research directions, aiming to highlight advancements and challenges in the field of anomaly or OOD detection and encourage further progress.

\section*{Limitations}
While this survey provides a comprehensive overview of the utilization of Large Language Models (LLMs) for anomaly and out-of-distribution (OOD) detection, several limitations should be acknowledged:
\begin{itemize}[leftmargin=*,itemsep=0.2pt,topsep=1.pt]
    \item \textbf{Scope of Coverage}: Although we endeavored to include the latest research, the rapid pace of advancements in the field means that some recent developments may not be covered.
    \item \textbf{Depth of Analysis}: Given the broad range of topics discussed, certain methods may not be explored in the depth they deserve.
    \item \textbf{Evaluations and Benchmarks}: Due to space constraints, we did not include a detailed summary of common evaluation metrics and benchmark datasets used in this area.
\end{itemize}
By acknowledging these limitations, we aim to provide a balanced perspective and encourage further research to address these gaps and build on the foundations laid by this survey.
\bibliography{anthology,custom}

\begin{thebibliography}{140}
\providecommand{\natexlab}[1]{#1}

\bibitem[{Abati et~al.(2019)Abati, Porrello, Calderara, and Cucchiara}]{autogres19cvpr}
Davide Abati, Angelo Porrello, Simone Calderara, and Rita Cucchiara. 2019.
\newblock Latent space autoregression for novelty detection.
\newblock In \emph{CVPR}.

\bibitem[{Achiam et~al.(2023)Achiam, Adler, Agarwal, Ahmad, Akkaya, Aleman, Almeida, Altenschmidt, Altman, Anadkat et~al.}]{achiam2023gpt}
Josh Achiam, Steven Adler, Sandhini Agarwal, Lama Ahmad, Ilge Akkaya, Florencia~Leoni Aleman, Diogo Almeida, Janko Altenschmidt, Sam Altman, Shyamal Anadkat, et~al. 2023.
\newblock Gpt-4 technical report.
\newblock \emph{arXiv preprint arXiv:2303.08774}.

\bibitem[{Alayrac et~al.(2022)Alayrac, Donahue, Luc, Miech, Barr, Hasson, Lenc, Mensch, Millican, Reynolds et~al.}]{alayrac2022flamingo}
Jean-Baptiste Alayrac, Jeff Donahue, Pauline Luc, Antoine Miech, Iain Barr, Yana Hasson, Karel Lenc, Arthur Mensch, Katherine Millican, Malcolm Reynolds, et~al. 2022.
\newblock Flamingo: a visual language model for few-shot learning.
\newblock \emph{NeurIPS}.

\bibitem[{Almodovar et~al.(2024)Almodovar, Sabrina, Karimi, and Azad}]{LogFiT}
Crispin Almodovar, Fariza Sabrina, Sarvnaz Karimi, and Salahuddin Azad. 2024.
\newblock \href {https://doi.org/10.1109/TNSM.2024.3358730} {Logfit: Log anomaly detection using fine-tuned language models}.
\newblock \emph{IEEE Transactions on Network and Service Management}.

\bibitem[{Alnegheimish et~al.(2024)Alnegheimish, Nguyen, Berti-Equille, and Veeramachaneni}]{alnegheimish2024large}
Sarah Alnegheimish, Linh Nguyen, Laure Berti-Equille, and Kalyan Veeramachaneni. 2024.
\newblock Large language models can be zero-shot anomaly detectors for time series?
\newblock \emph{arXiv preprint arXiv:2405.14755}.

\bibitem[{Bai et~al.(2024)Bai, Han, Cao, Jiang, Hu, and Zhang}]{bai2024id}
Yichen Bai, Zongbo Han, Bing Cao, Xiaoheng Jiang, Qinghua Hu, and Changqing Zhang. 2024.
\newblock \href {https://doi.org/10.48550/arXiv.2311.15243} {Id-like prompt learning for few-shot out-of-distribution detection}.
\newblock In \emph{CVPR}.

\bibitem[{Bejan et~al.(2023)Bejan, Manolache, and Popescu}]{bejan-etal-2023-ad}
Matei Bejan, Andrei Manolache, and Marius Popescu. 2023.
\newblock \href {https://doi.org/10.18653/v1/2023.emnlp-main.664} {{AD}-{NLP}: A benchmark for anomaly detection in natural language processing}.
\newblock In \emph{Proceedings of the 2023 Conference on Empirical Methods in Natural Language Processing}, pages 10766--10778, Singapore. Association for Computational Linguistics.

\bibitem[{Bekker and Davis(2020)}]{pusurvey20ml}
Jessa Bekker and Jesse Davis. 2020.
\newblock Learning from positive and unlabeled data: A survey.
\newblock \emph{Machine Learning}.

\bibitem[{Bengio et~al.(2013)Bengio, Courville, and Vincent}]{bengio2013representation}
Yoshua Bengio, Aaron Courville, and Pascal Vincent. 2013.
\newblock Representation learning: A review and new perspectives.
\newblock \emph{IEEE transactions on pattern analysis and machine intelligence}.

\bibitem[{Bergmann et~al.(2022)Bergmann, Batzner, Fauser, Sattlegger, and Steger}]{bergmann2022beyond}
Paul Bergmann, Kilian Batzner, Michael Fauser, David Sattlegger, and Carsten Steger. 2022.
\newblock Beyond dents and scratches: Logical constraints in unsupervised anomaly detection and localization.
\newblock \emph{IJCV}.

\bibitem[{Bergmann et~al.(2019)Bergmann, Fauser, Sattlegger, and Steger}]{8954181}
Paul Bergmann, Michael Fauser, David Sattlegger, and Carsten Steger. 2019.
\newblock Mvtec ad — a comprehensive real-world dataset for unsupervised anomaly detection.
\newblock In \emph{CVPR}.

\bibitem[{Brown et~al.(2020)Brown, Mann, Ryder, Subbiah, Kaplan, Dhariwal, Neelakantan, Shyam, Sastry, Askell et~al.}]{brown2020language}
Tom Brown, Benjamin Mann, Nick Ryder, Melanie Subbiah, Jared~D Kaplan, Prafulla Dhariwal, Arvind Neelakantan, Pranav Shyam, Girish Sastry, Amanda Askell, et~al. 2020.
\newblock Language models are few-shot learners.
\newblock \emph{NeurIPS}.

\bibitem[{Cao et~al.(2024)Cao, Zhong, Zhou, Liu, Liu, and Han}]{cao2024envisioning}
Chentao Cao, Zhun Zhong, Zhanke Zhou, Yang Liu, Tongliang Liu, and Bo~Han. 2024.
\newblock Envisioning outlier exposure by large language models for out-of-distribution detection.
\newblock In \emph{ICML}.

\bibitem[{Cao et~al.(2023)Cao, Xu, Sun, Huang, and Shen}]{cao2023towards}
Yunkang Cao, Xiaohao Xu, Chen Sun, Xiaonan Huang, and Weiming Shen. 2023.
\newblock Towards generic anomaly detection and understanding: Large-scale visual-linguistic model (gpt-4v) takes the lead.
\newblock \emph{arXiv preprint arXiv:2311.02782}.

\bibitem[{Chen et~al.(2024)Chen, Zhang, Zheng, and Wang}]{chen2024tagfog}
Jiankang Chen, Tong Zhang, Wei-Shi Zheng, and Ruixuan Wang. 2024.
\newblock Tagfog: Textual anchor guidance and fake outlier generation for visual out-of-distribution detection.
\newblock In \emph{AAAI}.

\bibitem[{Chen et~al.(2023{\natexlab{a}})Chen, Han, and Zhang}]{chen2023april}
Xuhai Chen, Yue Han, and Jiangning Zhang. 2023{\natexlab{a}}.
\newblock April-gan: A zero-/few-shot anomaly classification and segmentation method for cvpr 2023 vand workshop challenge tracks 1\&2: 1st place on zero-shot ad and 4th place on few-shot ad.
\newblock \emph{arXiv preprint arXiv:2305.17382}.

\bibitem[{Chen et~al.(2023{\natexlab{b}})Chen, Zhang, Tian, He, Zhang, Wang, Wang, Wu, and Liu}]{chen2023clip}
Xuhai Chen, Jiangning Zhang, Guanzhong Tian, Haoyang He, Wuhao Zhang, Yabiao Wang, Chengjie Wang, Yunsheng Wu, and Yong Liu. 2023{\natexlab{b}}.
\newblock Clip-ad: A language-guided staged dual-path model for zero-shot anomaly detection.
\newblock \emph{arXiv preprint arXiv:2311.00453}.

\bibitem[{Chowdhery et~al.(2023)Chowdhery, Narang, Devlin, Bosma, Mishra, Roberts, Barham, Chung, Sutton, Gehrmann et~al.}]{chowdhery2023palm}
Aakanksha Chowdhery, Sharan Narang, Jacob Devlin, Maarten Bosma, Gaurav Mishra, Adam Roberts, Paul Barham, Hyung~Won Chung, Charles Sutton, Sebastian Gehrmann, et~al. 2023.
\newblock Palm: Scaling language modeling with pathways.
\newblock \emph{JMLR}.

\bibitem[{Cimpoi et~al.(2014)Cimpoi, Maji, Kokkinos, Mohamed, and Vedaldi}]{texture}
Mircea Cimpoi, Subhransu Maji, Iasonas Kokkinos, Sammy Mohamed, and Andrea Vedaldi. 2014.
\newblock Describing textures in the wild.
\newblock In \emph{CVPR}.

\bibitem[{Dai et~al.(2023)Dai, Lang, Zeng, Huang, and Li}]{dai2023exploring}
Yi~Dai, Hao Lang, Kaisheng Zeng, Fei Huang, and Yongbin Li. 2023.
\newblock Exploring large language models for multi-modal out-of-distribution detection.
\newblock \emph{arXiv preprint arXiv:2310.08027}.

\bibitem[{Damm et~al.(2024)Damm, Laszkiewicz, Lederer, and Fischer}]{damm2024anomalydino}
Simon Damm, Mike Laszkiewicz, Johannes Lederer, and Asja Fischer. 2024.
\newblock Anomalydino: Boosting patch-based few-shot anomaly detection with dinov2.
\newblock \emph{arXiv preprint arXiv:2405.14529}.

\bibitem[{Danuser and Stricker(1998)}]{anomalyparametric98pami}
Gaudenz Danuser and Markus Stricker. 1998.
\newblock Parametric model fitting: From inlier characterization to outlier detection.
\newblock \emph{TPAMI}.

\bibitem[{Deecke et~al.(2018)Deecke, Vandermeulen, Ruff, Mandt, and Kloft}]{adgan18ecml}
Lucas Deecke, Robert Vandermeulen, Lukas Ruff, Stephan Mandt, and Marius Kloft. 2018.
\newblock Image anomaly detection with generative adversarial networks.
\newblock In \emph{ECML\&KDD}.

\bibitem[{Defard et~al.(2021)Defard, Setkov, Loesch, and Audigier}]{defard2021padim}
Thomas Defard, Aleksandr Setkov, Angelique Loesch, and Romaric Audigier. 2021.
\newblock Padim: a patch distribution modeling framework for anomaly detection and localization.
\newblock In \emph{International Conference on Pattern Recognition}, pages 475--489. Springer.

\bibitem[{Deng et~al.(2023)Deng, Zhang, Bao, and Li}]{deng2023anovl}
Hanqiu Deng, Zhaoxiang Zhang, Jinan Bao, and Xingyu Li. 2023.
\newblock Anovl: Adapting vision-language models for unified zero-shot anomaly localization.
\newblock \emph{arXiv preprint arXiv:2308.15939}.

\bibitem[{Deng et~al.(2009)Deng, Dong, Socher, Li, Li, and Fei-Fei}]{5206848}
Jia Deng, Wei Dong, Richard Socher, Li-Jia Li, Kai Li, and Li~Fei-Fei. 2009.
\newblock Imagenet: A large-scale hierarchical image database.
\newblock In \emph{CVPR}.

\bibitem[{Devlin et~al.(2018)Devlin, Chang, Lee, and Toutanova}]{devlin2018bert}
Jacob Devlin, Ming-Wei Chang, Kenton Lee, and Kristina Toutanova. 2018.
\newblock Bert: Pre-training of deep bidirectional transformers for language understanding.
\newblock \emph{arXiv preprint arXiv:1810.04805}.

\bibitem[{Ding et~al.(2024)Ding, Qin, Zhao, Luo, Li, Chen, Xia, Hu, Luu, and Joty}]{ding-etal-2024-data}
Bosheng Ding, Chengwei Qin, Ruochen Zhao, Tianze Luo, Xinze Li, Guizhen Chen, Wenhan Xia, Junjie Hu, Anh~Tuan Luu, and Shafiq Joty. 2024.
\newblock \href {https://doi.org/10.18653/v1/2024.findings-acl.97} {Data augmentation using {LLM}s: Data perspectives, learning paradigms and challenges}.
\newblock In \emph{Findings of the Association for Computational Linguistics: ACL 2024}, pages 1679--1705, Bangkok, Thailand. Association for Computational Linguistics.

\bibitem[{Elhafsi et~al.(2023)Elhafsi, Sinha, Agia, Schmerling, Nesnas, and Pavone}]{elhafsi2023semantic}
Amine Elhafsi, Rohan Sinha, Christopher Agia, Edward Schmerling, Issa~AD Nesnas, and Marco Pavone. 2023.
\newblock \href {https://doi.org/10.1007/s10514-023-10132-6} {Semantic anomaly detection with large language models}.
\newblock \emph{Autonomous Robots}.

\bibitem[{Esmaeilpour et~al.(2022)Esmaeilpour, Liu, Robertson, and Shu}]{esmaeilpour2022zero}
Sepideh Esmaeilpour, Bing Liu, Eric Robertson, and Lei Shu. 2022.
\newblock Zero-shot out-of-distribution detection based on the pre-trained model clip.
\newblock In \emph{AAAI}.

\bibitem[{Felipe(2018)}]{Kitamura2018headct}
Kitamura Felipe. 2018.
\newblock Head ct - hemorrhage.
\newblock \url{https://www. kaggle.com/datasets/felipekitamura/headct-hemorrhage}.

\bibitem[{Fort et~al.(2021)Fort, Ren, and Lakshminarayanan}]{fort2021exploring}
Stanislav Fort, Jie Ren, and Balaji Lakshminarayanan. 2021.
\newblock Exploring the limits of out-of-distribution detection.
\newblock In \emph{NeurIPS}.

\bibitem[{Fu et~al.(2024)Fu, Patel, Krishnamurthy, and Khorrami}]{fu2024clipscope}
Hao Fu, Naman Patel, Prashanth Krishnamurthy, and Farshad Khorrami. 2024.
\newblock Clipscope: Enhancing zero-shot ood detection with bayesian scoring.
\newblock \emph{arXiv preprint arXiv:2405.14737}.

\bibitem[{Goodfellow et~al.(2014)Goodfellow, Pouget-Abadie, Mirza, Xu, Warde-Farley, Ozair, Courville, and Bengio}]{gan14nips}
Ian Goodfellow, Jean Pouget-Abadie, Mehdi Mirza, Bing Xu, David Warde-Farley, Sherjil Ozair, Aaron Courville, and Yoshua Bengio. 2014.
\newblock Generative adversarial nets.
\newblock In \emph{NeurIPS}.

\bibitem[{Gu et~al.(2024{\natexlab{a}})Gu, Zhu, Zhu, Chen, Li, Tang, and Wang}]{gu2024filo}
Zhaopeng Gu, Bingke Zhu, Guibo Zhu, Yingying Chen, Hao Li, Ming Tang, and Jinqiao Wang. 2024{\natexlab{a}}.
\newblock Filo: Zero-shot anomaly detection by fine-grained description and high-quality localization.
\newblock \emph{arXiv preprint arXiv:2404.13671}.

\bibitem[{Gu et~al.(2024{\natexlab{b}})Gu, Zhu, Zhu, Chen, Tang, and Wang}]{gu2024anomalygpt}
Zhaopeng Gu, Bingke Zhu, Guibo Zhu, Yingying Chen, Ming Tang, and Jinqiao Wang. 2024{\natexlab{b}}.
\newblock Anomalygpt: Detecting industrial anomalies using large vision-language models.
\newblock In \emph{AAAI}.

\bibitem[{Guidotti et~al.(2019)Guidotti, Monreale, Ruggieri, Turini, Giannotti, and Pedreschi}]{guidotti2019survey}
Riccardo Guidotti, Anna Monreale, Salvatore Ruggieri, Franco Turini, Fosca Giannotti, and Dino Pedreschi. 2019.
\newblock A survey of methods for explaining black box models.
\newblock \emph{ACM computing surveys (CSUR)}.

\bibitem[{Hadadi et~al.(2024)Hadadi, Xu, Bianculli, and Briand}]{hadadi2024anomaly}
Fatemeh Hadadi, Qinghua Xu, Domenico Bianculli, and Lionel Briand. 2024.
\newblock Anomaly detection on unstable logs with gpt models.
\newblock \emph{arXiv preprint arXiv:2406.07467}.

\bibitem[{Hendrycks et~al.(2019)Hendrycks, Basart, Mazeika, Zou, Kwon, Mostajabi, Steinhardt, and Song}]{hendrycks2019scaling}
Dan Hendrycks, Steven Basart, Mantas Mazeika, Andy Zou, Joe Kwon, Mohammadreza Mostajabi, Jacob Steinhardt, and Dawn Song. 2019.
\newblock Scaling out-of-distribution detection for real-world settings.
\newblock \emph{arXiv preprint arXiv:1911.11132}.

\bibitem[{Hendrycks and Gimpel(2017)}]{msp17iclr}
Dan Hendrycks and Kevin Gimpel. 2017.
\newblock A baseline for detecting misclassified and out-of-distribution examples in neural networks.
\newblock In \emph{ICLR}.

\bibitem[{Hilal et~al.(2022)Hilal, Gadsden, and Yawney}]{HILAL2022116429}
Waleed Hilal, S.~Andrew Gadsden, and John Yawney. 2022.
\newblock \href {https://doi.org/10.1016/j.eswa.2021.116429} {Financial fraud: A review of anomaly detection techniques and recent advances}.
\newblock \emph{Expert Systems with Applications}.

\bibitem[{Holzinger et~al.(2019)Holzinger, Langs, Denk, Zatloukal, and M{\"u}ller}]{holzinger2019causability}
Andreas Holzinger, Georg Langs, Daniel Denk, Kurt Zatloukal, and Henning M{\"u}ller. 2019.
\newblock Causability and explainability of artificial intelligence in medicine.
\newblock \emph{Wiley Interdisciplinary Reviews: Data Mining and Knowledge Discovery}.

\bibitem[{Hu et~al.(2018)Hu, Gao, Li, Wu, Du, and Maybank}]{anomalykde18tkde}
Weiming Hu, Jun Gao, Bing Li, Ou~Wu, Junping Du, and Stephen Maybank. 2018.
\newblock Anomaly detection using local kernel density estimation and context-based regression.
\newblock \emph{TKDE}.

\bibitem[{Hu et~al.(2023)Hu, Wang, Lan, Xu, Lim, Bing, Xu, Poria, and Lee}]{hu2023llm}
Zhiqiang Hu, Lei Wang, Yihuai Lan, Wanyu Xu, Ee-Peng Lim, Lidong Bing, Xing Xu, Soujanya Poria, and Roy Ka-Wei Lee. 2023.
\newblock Llm-adapters: An adapter family for parameter-efficient fine-tuning of large language models.
\newblock \emph{arXiv preprint arXiv:2304.01933}.

\bibitem[{Huang et~al.(2024{\natexlab{a}})Huang, Jiang, Feng, Zhang, Wang, and Wang}]{huang2024adapting}
Chaoqin Huang, Aofan Jiang, Jinghao Feng, Ya~Zhang, Xinchao Wang, and Yanfeng Wang. 2024{\natexlab{a}}.
\newblock Adapting visual-language models for generalizable anomaly detection in medical images.
\newblock In \emph{CVPR}.

\bibitem[{Huang et~al.(2024{\natexlab{b}})Huang, Song, Su, and Wang}]{huang2024out}
K~Huang, G~Song, Hanwen Su, and Jiyan Wang. 2024{\natexlab{b}}.
\newblock Out-of-distribution detection using peer-class generated by large language model.
\newblock \emph{arXiv preprint arXiv:2403.13324}.

\bibitem[{Jaskie and Spanias(2019)}]{pusurvey19iisa}
Kristen Jaskie and Andreas Spanias. 2019.
\newblock Positive and unlabeled learning algorithms and applications: A survey.
\newblock In \emph{International Conference on Information, Intelligence, Systems and Applications}.

\bibitem[{Jeong et~al.(2023)Jeong, Zou, Kim, Zhang, Ravichandran, and Dabeer}]{jeong2023winclip}
Jongheon Jeong, Yang Zou, Taewan Kim, Dongqing Zhang, Avinash Ravichandran, and Onkar Dabeer. 2023.
\newblock Winclip: Zero-/few-shot anomaly classification and segmentation.
\newblock In \emph{CVPR}.

\bibitem[{Jiang et~al.(2024)Jiang, Liu, Fang, Chen, Liu, Zheng, and Han}]{jiang2024negative}
Xue Jiang, Feng Liu, Zhen Fang, Hong Chen, Tongliang Liu, Feng Zheng, and Bo~Han. 2024.
\newblock Negative label guided ood detection with pretrained vision-language models.
\newblock \emph{arXiv preprint arXiv:2403.20078}.

\bibitem[{Kermany et~al.(2018)Kermany, Goldbaum, Cai, Valentim, Liang, Baxter, McKeown, Yang, Wu, Yan et~al.}]{kermany2018identifying}
Daniel~S Kermany, Michael Goldbaum, Wenjia Cai, Carolina~CS Valentim, Huiying Liang, Sally~L Baxter, Alex McKeown, Ge~Yang, Xiaokang Wu, Fangbing Yan, et~al. 2018.
\newblock Identifying medical diagnoses and treatable diseases by image-based deep learning.
\newblock \emph{Cell}.

\bibitem[{Kingma and Dhariwal(2018)}]{glow18nips}
Diederik~P Kingma and Prafulla Dhariwal. 2018.
\newblock Glow: Generative flow with invertible 1x1 convolutions.
\newblock \emph{NeurIPS}.

\bibitem[{Kingma and Welling(2013)}]{vae13arxiv}
Diederik~P Kingma and Max Welling. 2013.
\newblock Auto-encoding variational bayes.
\newblock \emph{arXiv preprint arXiv:1312.6114}.

\bibitem[{Kobyzev et~al.(2020)Kobyzev, Prince, and Brubaker}]{flowreview20pami}
Ivan Kobyzev, Simon Prince, and Marcus Brubaker. 2020.
\newblock Normalizing flows: An introduction and review of current methods.
\newblock \emph{TPAMI}.

\bibitem[{Kojima et~al.(2022)Kojima, Gu, Reid, Matsuo, and Iwasawa}]{kojima2022large}
Takeshi Kojima, Shixiang~Shane Gu, Machel Reid, Yutaka Matsuo, and Yusuke Iwasawa. 2022.
\newblock Large language models are zero-shot reasoners.
\newblock \emph{NeurIPS}.

\bibitem[{Kramer(1991)}]{ae91aiche}
Mark~A Kramer. 1991.
\newblock Nonlinear principal component analysis using autoassociative neural networks.
\newblock \emph{AIChE journal}.

\bibitem[{Krizhevsky et~al.(2012)Krizhevsky, Sutskever, and Hinton}]{closedset}
Alex Krizhevsky, Ilya Sutskever, and Geoffrey~E. Hinton. 2012.
\newblock \href {https://api.semanticscholar.org/CorpusID:195908774} {Imagenet classification with deep convolutional neural networks}.
\newblock \emph{Communications of the ACM}.

\bibitem[{Lafon et~al.(2024)Lafon, Ramzi, Rambour, Audebert, and Thome}]{lafon2024gallop}
Marc Lafon, Elias Ramzi, Cl{\'e}ment Rambour, Nicolas Audebert, and Nicolas Thome. 2024.
\newblock Gallop: Learning global and local prompts for vision-language models.
\newblock \emph{arXiv preprint arXiv:2407.01400}.

\bibitem[{Lang(1995)}]{LANG1995331}
Ken Lang. 1995.
\newblock Newsweeder: Learning to filter netnews.
\newblock In \emph{Machine Learning Proceedings 1995}. Morgan Kaufmann, San Francisco (CA).

\bibitem[{Lee et~al.(2018)Lee, Lee, Lee, and Shin}]{lee2018simple}
Kimin Lee, Kibok Lee, Honglak Lee, and Jinwoo Shin. 2018.
\newblock A simple unified framework for detecting out-of-distribution samples and adversarial attacks.
\newblock In \emph{NeurIPS}.

\bibitem[{Lewis et~al.(2019)Lewis, Liu, Goyal, Ghazvininejad, Mohamed, Levy, Stoyanov, and Zettlemoyer}]{lewis2019bart}
Mike Lewis, Yinhan Liu, Naman Goyal, Marjan Ghazvininejad, Abdelrahman Mohamed, Omer Levy, Veselin Stoyanov, and Luke Zettlemoyer. 2019.
\newblock Bart: Denoising sequence-to-sequence pre-training for natural language generation, translation, and comprehension.
\newblock \emph{arXiv preprint arXiv:1910.13461}.

\bibitem[{Leys et~al.(2018)Leys, Klein, Dominicy, and Ley}]{mahalanobis18jesp}
Christophe Leys, Olivier Klein, Yves Dominicy, and Christophe Ley. 2018.
\newblock Detecting multivariate outliers: Use a robust variant of the mahalanobis distance.
\newblock \emph{Journal of Experimental Social Psychology}.

\bibitem[{Li et~al.(2024{\natexlab{a}})Li, Zhao, Qiu, Kloft, Smyth, Rudolph, and Mandt}]{li2024anomaly}
Aodong Li, Yunhan Zhao, Chen Qiu, Marius Kloft, Padhraic Smyth, Maja Rudolph, and Stephan Mandt. 2024{\natexlab{a}}.
\newblock Anomaly detection of tabular data using llms.
\newblock \emph{arXiv preprint arXiv:2406.16308}.

\bibitem[{Li et~al.(2023{\natexlab{a}})Li, Li, Savarese, and Fei-Fei}]{li2023blip}
Junnan Li, Dongxu Li, Silvio Savarese, and Li~Fei-Fei. 2023{\natexlab{a}}.
\newblock Blip-2: Bootstrapping language-image pre-training with frozen image encoders and large language models.
\newblock \emph{arXiv preprint arXiv:2301.12597}.

\bibitem[{Li et~al.(2024{\natexlab{b}})Li, Pang, Bai, Miao, and Zheng}]{li2024learning}
Tianqi Li, Guansong Pang, Xiao Bai, Wenjun Miao, and Jin Zheng. 2024{\natexlab{b}}.
\newblock Learning transferable negative prompts for out-of-distribution detection.
\newblock In \emph{CVPR}.

\bibitem[{Li et~al.(2024{\natexlab{c}})Li, Ding, Wang, and Lee}]{li-etal-2024-empowering}
Yichuan Li, Kaize Ding, Jianling Wang, and Kyumin Lee. 2024{\natexlab{c}}.
\newblock \href {https://doi.org/10.18653/v1/2024.findings-acl.756} {Empowering large language models for textual data augmentation}.
\newblock In \emph{Findings of the Association for Computational Linguistics: ACL 2024}, pages 12734--12751, Bangkok, Thailand. Association for Computational Linguistics.

\bibitem[{Li et~al.(2024{\natexlab{d}})Li, Xiong, Chen, and Chen}]{li2024setar}
Yixia Li, Boya Xiong, Guanhua Chen, and Yun Chen. 2024{\natexlab{d}}.
\newblock Setar: Out-of-distribution detection with selective low-rank approximation.
\newblock \emph{arXiv preprint arXiv:2406.12629}.

\bibitem[{Li et~al.(2024{\natexlab{e}})Li, Li, Xiao, Yang, Nian, Hu, and Zhao}]{li2024nlp}
Yuangang Li, Jiaqi Li, Zhuo Xiao, Tiankai Yang, Yi~Nian, Xiyang Hu, and Yue Zhao. 2024{\natexlab{e}}.
\newblock Nlp-adbench: Nlp anomaly detection benchmark.
\newblock \emph{arXiv preprint arXiv:2412.04784}.

\bibitem[{Li et~al.(2023{\natexlab{b}})Li, Wang, Yuan, Liu, Zhao, Guo, Xu, Shi, and Zuo}]{li2023myriad}
Yuanze Li, Haolin Wang, Shihao Yuan, Ming Liu, Debin Zhao, Yiwen Guo, Chen Xu, Guangming Shi, and Wangmeng Zuo. 2023{\natexlab{b}}.
\newblock Myriad: Large multimodal model by applying vision experts for industrial anomaly detection.
\newblock \emph{arXiv preprint arXiv:2310.19070}.

\bibitem[{Liang et~al.(2018)Liang, Li, and Srikant}]{liang2018enhancing}
Shiyu Liang, Yixuan Li, and R.~Srikant. 2018.
\newblock Enhancing the reliability of out-of-distribution image detection in neural networks.
\newblock In \emph{International Conference on Learning Representations}.

\bibitem[{Liu et~al.(2024{\natexlab{a}})Liu, Zhan, Lu, Feng, Xue, and Wu}]{liu-etal-2024-good}
Bo~Liu, Li-Ming Zhan, Zexin Lu, Yujie Feng, Lei Xue, and Xiao-Ming Wu. 2024{\natexlab{a}}.
\newblock \href {https://aclanthology.org/2024.lrec-main.720/} {How good are {LLM}s at out-of-distribution detection?}
\newblock In \emph{Proceedings of the 2024 Joint International Conference on Computational Linguistics, Language Resources and Evaluation (LREC-COLING 2024)}, pages 8211--8222, Torino, Italia. ELRA and ICCL.

\bibitem[{Liu et~al.(2024{\natexlab{b}})Liu, Xie, Wang, Li, Wang, Zheng, and Jin}]{liu2024deep}
Jiaqi Liu, Guoyang Xie, Jinbao Wang, Shangnian Li, Chengjie Wang, Feng Zheng, and Yaochu Jin. 2024{\natexlab{b}}.
\newblock Deep industrial image anomaly detection: A survey.
\newblock \emph{Machine Intelligence Research}.

\bibitem[{Liu et~al.(2024{\natexlab{c}})Liu, Zhang, Qian, Ma, Qin, Bansal, Lin, Rajmohan, and Zhang}]{liu2024large}
Jun Liu, Chaoyun Zhang, Jiaxu Qian, Minghua Ma, Si~Qin, Chetan Bansal, Qingwei Lin, Saravan Rajmohan, and Dongmei Zhang. 2024{\natexlab{c}}.
\newblock Large language models can deliver accurate and interpretable time series anomaly detection.
\newblock \emph{arXiv preprint arXiv:2405.15370}.

\bibitem[{Liu et~al.(2020)Liu, Wang, Owens, and Li}]{liu2020energy}
Weitang Liu, Xiaoyun Wang, John Owens, and Yixuan Li. 2020.
\newblock Energy-based out-of-distribution detection.
\newblock \emph{NeurIPS}.

\bibitem[{Liu et~al.(2024{\natexlab{d}})Liu, Tao, Meng, Yao, Zhao, and Yang}]{liu2024logprompt}
Yilun Liu, Shimin Tao, Weibin Meng, Feiyu Yao, Xiaofeng Zhao, and Hao Yang. 2024{\natexlab{d}}.
\newblock Logprompt: Prompt engineering towards zero-shot and interpretable log analysis.
\newblock In \emph{ICSE}.

\bibitem[{Liu et~al.(2019)Liu, Ott, Goyal, Du, Joshi, Chen, Levy, Lewis, Zettlemoyer, and Stoyanov}]{liu2019roberta}
Yinhan Liu, Myle Ott, Naman Goyal, Jingfei Du, Mandar Joshi, Danqi Chen, Omer Levy, Mike Lewis, Luke Zettlemoyer, and Veselin Stoyanov. 2019.
\newblock Roberta: A robustly optimized bert pretraining approach.
\newblock \emph{arXiv preprint arXiv:1907.11692}.

\bibitem[{Liznerski et~al.(2022)Liznerski, Ruff, Vandermeulen, Franks, M{\"u}ller, and Kloft}]{liznerski2022exposing}
Philipp Liznerski, Lukas Ruff, Robert~A Vandermeulen, Billy~Joe Franks, Klaus-Robert M{\"u}ller, and Marius Kloft. 2022.
\newblock Exposing outlier exposure: What can be learned from few, one, and zero outlier images.
\newblock \emph{arXiv preprint arXiv:2205.11474}.

\bibitem[{Lv and Sun(2024)}]{lv2024video}
Hui Lv and Qianru Sun. 2024.
\newblock Video anomaly detection and explanation via large language models.
\newblock \emph{arXiv preprint arXiv:2401.05702}.

\bibitem[{Ming et~al.(2022)Ming, Cai, Gu, Sun, Li, and Li}]{ming2022delving}
Yifei Ming, Ziyang Cai, Jiuxiang Gu, Yiyou Sun, Wei Li, and Yixuan Li. 2022.
\newblock Delving into out-of-distribution detection with vision-language representations.
\newblock In \emph{NeurIPS}.

\bibitem[{Ming and Li(2024)}]{ming2024does}
Yifei Ming and Yixuan Li. 2024.
\newblock How does fine-tuning impact out-of-distribution detection for vision-language models?
\newblock \emph{IJCV}.

\bibitem[{Miyai et~al.(2024{\natexlab{a}})Miyai, Yang, Zhang, Ming, Lin, Yu, Irie, Joty, Li, Li et~al.}]{miyai2024generalized}
Atsuyuki Miyai, Jingkang Yang, Jingyang Zhang, Yifei Ming, Yueqian Lin, Qing Yu, Go~Irie, Shafiq Joty, Yixuan Li, Hai Li, et~al. 2024{\natexlab{a}}.
\newblock Generalized out-of-distribution detection and beyond in vision language model era: A survey.
\newblock \emph{arXiv preprint arXiv:2407.21794}.

\bibitem[{Miyai et~al.(2024{\natexlab{b}})Miyai, Yang, Zhang, Ming, Yu, Irie, Li, Li, Liu, and Aizawa}]{miyai2024unsolvable}
Atsuyuki Miyai, Jingkang Yang, Jingyang Zhang, Yifei Ming, Qing Yu, Go~Irie, Yixuan Li, Hai Li, Ziwei Liu, and Kiyoharu Aizawa. 2024{\natexlab{b}}.
\newblock Unsolvable problem detection: Evaluating trustworthiness of vision language models.
\newblock \emph{arXiv preprint arXiv:2403.20331}.

\bibitem[{Miyai et~al.(2023)Miyai, Yu, Irie, and Aizawa}]{miyai2023zero}
Atsuyuki Miyai, Qing Yu, Go~Irie, and Kiyoharu Aizawa. 2023.
\newblock Zero-shot in-distribution detection in multi-object settings using vision-language foundation models.
\newblock \emph{arXiv preprint arXiv:2304.04521}.

\bibitem[{Miyai et~al.(2024{\natexlab{c}})Miyai, Yu, Irie, and Aizawa}]{miyai2024locoop}
Atsuyuki Miyai, Qing Yu, Go~Irie, and Kiyoharu Aizawa. 2024{\natexlab{c}}.
\newblock Locoop: Few-shot out-of-distribution detection via prompt learning.
\newblock \emph{NeurIPS}.

\bibitem[{Nie et~al.(2024)Nie, Zhang, Fang, Liu, Han, and Tian}]{nie2024outofdistribution}
Jun Nie, Yonggang Zhang, Zhen Fang, Tongliang Liu, Bo~Han, and Xinmei Tian. 2024.
\newblock Out-of-distribution detection with negative prompts.
\newblock In \emph{ICLR}.

\bibitem[{OpenAI(2023)}]{openai2023}
OpenAI. 2023.
\newblock Gpt-4 technical report.
\newblock \emph{arXiv preprint arXiv:2303.08774}.

\bibitem[{Pang et~al.(2021)Pang, Shen, Cao, and Hengel}]{pang2021deep}
Guansong Pang, Chunhua Shen, Longbing Cao, and Anton Van~Den Hengel. 2021.
\newblock Deep learning for anomaly detection: A review.
\newblock \emph{ACM computing surveys (CSUR)}.

\bibitem[{Paparrizos et~al.(2022)Paparrizos, Kang, Boniol, Tsay, Palpanas, and Franklin}]{10.14778/3529337.3529354}
John Paparrizos, Yuhao Kang, Paul Boniol, Ruey~S. Tsay, Themis Palpanas, and Michael~J. Franklin. 2022.
\newblock \href {https://doi.org/10.14778/3529337.3529354} {Tsb-uad: an end-to-end benchmark suite for univariate time-series anomaly detection}.
\newblock \emph{Proc. VLDB Endow.}

\bibitem[{Park et~al.(2023)Park, Mok, Jung, Lee, and Yoon}]{park2023on}
Sangha Park, Jisoo Mok, Dahuin Jung, Saehyung Lee, and Sungroh Yoon. 2023.
\newblock On the powerfulness of textual outlier exposure for visual ood detection.
\newblock In \emph{NeurIPS}.

\bibitem[{Park et~al.(2019)Park, Liu, Wang, and Zhu}]{park2019semantic}
Taesung Park, Ming-Yu Liu, Ting-Chun Wang, and Jun-Yan Zhu. 2019.
\newblock Semantic image synthesis with spatially-adaptive normalization.
\newblock In \emph{CVPR}.

\bibitem[{Parzen(1962)}]{kde62math}
Emanuel Parzen. 1962.
\newblock On estimation of a probability density function and mode.
\newblock \emph{The annals of mathematical statistics}.

\bibitem[{Pidhorskyi et~al.(2018)Pidhorskyi, Almohsen, Adjeroh, and Doretto}]{gpnd18nips}
Stanislav Pidhorskyi, Ranya Almohsen, Donald~A Adjeroh, and Gianfranco Doretto. 2018.
\newblock Generative probabilistic novelty detection with adversarial autoencoders.
\newblock In \emph{NeurIPS}.

\bibitem[{Qi et~al.(2023)Qi, Huang, Luan, Yang, Fung, Yang, Qian, Shang, Xiao, and Wu}]{qi2023loggpt}
Jiaxing Qi, Shaohan Huang, Zhongzhi Luan, Shu Yang, Carol Fung, Hailong Yang, Depei Qian, Jing Shang, Zhiwen Xiao, and Zhihui Wu. 2023.
\newblock \href {https://doi.org/10.1109/HPCC-DSS-SmartCity-DependSys60770.2023.00045} {Loggpt: Exploring chatgpt for log-based anomaly detection}.
\newblock In \emph{HPCC/DSS/SmartCity/DependSys}.

\bibitem[{Radford et~al.(2021)Radford, Kim, Hallacy, Ramesh, Goh, Agarwal, Sastry, Askell, Mishkin, Clark et~al.}]{radford2021learning}
Alec Radford, Jong~Wook Kim, Chris Hallacy, Aditya Ramesh, Gabriel Goh, Sandhini Agarwal, Girish Sastry, Amanda Askell, Pamela Mishkin, Jack Clark, et~al. 2021.
\newblock Learning transferable visual models from natural language supervision.
\newblock In \emph{ICML}.

\bibitem[{Raffel et~al.(2020)Raffel, Shazeer, Roberts, Lee, Narang, Matena, Zhou, Li, and Liu}]{raffel2020exploring}
Colin Raffel, Noam Shazeer, Adam Roberts, Katherine Lee, Sharan Narang, Michael Matena, Yanqi Zhou, Wei Li, and Peter~J Liu. 2020.
\newblock Exploring the limits of transfer learning with a unified text-to-text transformer.
\newblock \emph{JMLR}.

\bibitem[{Rayana(2016)}]{Rayana:2016}
Shebuti Rayana. 2016.
\newblock Odds library.

\bibitem[{Repository(2020)}]{clinc150_570}
UCI Machine~Learning Repository. 2020.
\newblock {CLINC150}.

\bibitem[{Roth et~al.(2022)Roth, Pemula, Zepeda, Sch{\"o}lkopf, Brox, and Gehler}]{roth2022towards}
Karsten Roth, Latha Pemula, Joaquin Zepeda, Bernhard Sch{\"o}lkopf, Thomas Brox, and Peter Gehler. 2022.
\newblock Towards total recall in industrial anomaly detection.
\newblock In \emph{CVPR}.

\bibitem[{Ruff et~al.(2018)Ruff, Vandermeulen, Goernitz, Deecke, Siddiqui, Binder, M{\"u}ller, and Kloft}]{deepsvdd18icml}
Lukas Ruff, Robert Vandermeulen, Nico Goernitz, Lucas Deecke, Shoaib~Ahmed Siddiqui, Alexander Binder, Emmanuel M{\"u}ller, and Marius Kloft. 2018.
\newblock Deep one-class classification.
\newblock In \emph{ICML}.

\bibitem[{Sabokrou et~al.(2018)Sabokrou, Khalooei, Fathy, and Adeli}]{advrec18cvpr}
Mohammad Sabokrou, Mohammad Khalooei, Mahmood Fathy, and Ehsan Adeli. 2018.
\newblock Adversarially learned one-class classifier for novelty detection.
\newblock In \emph{CVPR}.

\bibitem[{Sahoo et~al.(2024)Sahoo, Singh, Saha, Jain, Mondal, and Chadha}]{sahoo2024systematic}
Pranab Sahoo, Ayush~Kumar Singh, Sriparna Saha, Vinija Jain, Samrat Mondal, and Aman Chadha. 2024.
\newblock A systematic survey of prompt engineering in large language models: Techniques and applications.
\newblock \emph{arXiv preprint arXiv:2402.07927}.

\bibitem[{Salehi et~al.(2021)Salehi, Mirzaei, Hendrycks, Li, Rohban, and Sabokrou}]{salehi2021unified}
Mohammadreza Salehi, Hossein Mirzaei, Dan Hendrycks, Yixuan Li, Mohammad~Hossein Rohban, and Mohammad Sabokrou. 2021.
\newblock A unified survey on anomaly, novelty, open-set, and out-of-distribution detection: Solutions and future challenges.
\newblock \emph{arXiv preprint arXiv:2110.14051}.

\bibitem[{Sinha et~al.(2024)Sinha, Elhafsi, Agia, Foutter, Schmerling, and Pavone}]{sinha2024real}
Rohan Sinha, Amine Elhafsi, Christopher Agia, Matthew Foutter, Edward Schmerling, and Marco Pavone. 2024.
\newblock Real-time anomaly detection and reactive planning with large language models.
\newblock \emph{arXiv preprint arXiv:2407.08735}.

\bibitem[{Socher et~al.(2013)Socher, Perelygin, Wu, Chuang, Manning, Ng, and Potts}]{socher-etal-2013-recursive}
Richard Socher, Alex Perelygin, Jean Wu, Jason Chuang, Christopher~D. Manning, Andrew Ng, and Christopher Potts. 2013.
\newblock \href {https://aclanthology.org/D13-1170/} {Recursive deep models for semantic compositionality over a sentiment treebank}.
\newblock In \emph{Proceedings of the 2013 Conference on Empirical Methods in Natural Language Processing}, pages 1631--1642, Seattle, Washington, USA. Association for Computational Linguistics.

\bibitem[{Su et~al.(2024)Su, Jiang, Jin, Qiao, Xiao, Ma, Wei, Jing, Xu, and Lin}]{su2024large}
Jing Su, Chufeng Jiang, Xin Jin, Yuxin Qiao, Tingsong Xiao, Hongda Ma, Rong Wei, Zhi Jing, Jiajun Xu, and Junhong Lin. 2024.
\newblock Large language models for forecasting and anomaly detection: A systematic literature review.
\newblock \emph{arXiv preprint arXiv:2402.10350}.

\bibitem[{Sultani et~al.(2018)Sultani, Chen, and Shah}]{sultani2018real}
Waqas Sultani, Chen Chen, and Mubarak Shah. 2018.
\newblock Real-world anomaly detection in surveillance videos.
\newblock In \emph{CVPR}.

\bibitem[{Sun et~al.(2021)Sun, Guo, and Li}]{sun2021react}
Yiyou Sun, Chuan Guo, and Yixuan Li. 2021.
\newblock React: Out-of-distribution detection with rectified activations.
\newblock In \emph{NeurIPS}.

\bibitem[{Sun et~al.(2022)Sun, Ming, Zhu, and Li}]{sun2022knn}
Yiyou Sun, Yifei Ming, Xiaojin Zhu, and Yixuan Li. 2022.
\newblock Out-of-distribution detection with deep nearest neighbors.
\newblock In \emph{ICML}.

\bibitem[{Tack et~al.(2020)Tack, Mo, Jeong, and Shin}]{csi20nips}
Jihoon Tack, Sangwoo Mo, Jongheon Jeong, and Jinwoo Shin. 2020.
\newblock Csi: Novelty detection via contrastive learning on distributionally shifted instances.
\newblock In \emph{NeurIPS}.

\bibitem[{Tax(2002)}]{occ02tax}
David Martinus~Johannes Tax. 2002.
\newblock One-class classification: Concept learning in the absence of counter-examples.

\bibitem[{Team et~al.(2023)Team, Anil, Borgeaud, Wu, Alayrac, Yu, Soricut, Schalkwyk, Dai, Hauth et~al.}]{team2023gemini}
Gemini Team, Rohan Anil, Sebastian Borgeaud, Yonghui Wu, Jean-Baptiste Alayrac, Jiahui Yu, Radu Soricut, Johan Schalkwyk, Andrew~M Dai, Anja Hauth, et~al. 2023.
\newblock Gemini: a family of highly capable multimodal models.
\newblock \emph{arXiv preprint arXiv:2312.11805}.

\bibitem[{Tian et~al.(2014)Tian, Azarian, and Pecht}]{tian2014anomaly}
Jing Tian, Michael~H Azarian, and Michael Pecht. 2014.
\newblock Anomaly detection using self-organizing maps-based k-nearest neighbor algorithm.
\newblock In \emph{PHM Society European Conference}.

\bibitem[{Touvron et~al.(2023)Touvron, Lavril, Izacard, Martinet, Lachaux, Lacroix, Rozière, Goyal, Hambro, Azhar et~al.}]{touvron2023a}
Hugo Touvron, Thibaut Lavril, Gautier Izacard, Xavier Martinet, Marie-Anne Lachaux, Thomas Lacroix, Baptiste Rozière, Naman Goyal, Eric Hambro, Ferhan Azhar, et~al. 2023.
\newblock Llama: Open and efficient foundation language models.
\newblock \emph{arXiv preprint arXiv:2302.13971}.

\bibitem[{Turcotte et~al.(2016)Turcotte, Moore, Heard, and McPhall}]{poisson16isi}
Melissa Turcotte, Juston Moore, Nick Heard, and Aaron McPhall. 2016.
\newblock Poisson factorization for peer-based anomaly detection.
\newblock In \emph{IEEE Conference on Intelligence and Security Informatics (ISI)}.

\bibitem[{Van~Horn et~al.(2018)Van~Horn, Mac~Aodha, Song, Cui, Sun, Shepard, Adam, Perona, and Belongie}]{van2018inaturalist}
Grant Van~Horn, Oisin Mac~Aodha, Yang Song, Yin Cui, Chen Sun, Alex Shepard, Hartwig Adam, Pietro Perona, and Serge Belongie. 2018.
\newblock The inaturalist species classification and detection dataset.
\newblock In \emph{CVPR}.

\bibitem[{Van~Oord et~al.(2016)Van~Oord, Kalchbrenner, and Kavukcuoglu}]{pixelcnn16icml}
Aaron Van~Oord, Nal Kalchbrenner, and Koray Kavukcuoglu. 2016.
\newblock Pixel recurrent neural networks.
\newblock In \emph{ICML}.

\bibitem[{Wang et~al.(2023)Wang, Li, Yao, and Li}]{wang2023clipn}
Hualiang Wang, Yi~Li, Huifeng Yao, and Xiaomeng Li. 2023.
\newblock Clipn for zero-shot ood detection: Teaching clip to say no.
\newblock In \emph{ICCV}.

\bibitem[{Wang et~al.(2024)Wang, Chen, Chen, Wu, Zhu, Zeng, Luo, Lu, Zhou, Qiao et~al.}]{wang2024visionllm}
Wenhai Wang, Zhe Chen, Xiaokang Chen, Jiannan Wu, Xizhou Zhu, Gang Zeng, Ping Luo, Tong Lu, Jie Zhou, Yu~Qiao, et~al. 2024.
\newblock Visionllm: Large language model is also an open-ended decoder for vision-centric tasks.
\newblock In \emph{NeurIPS}.

\bibitem[{Wei et~al.(2022)Wei, Wang, Schuurmans, Bosma, Xia, Chi, Le, Zhou et~al.}]{wei2022chain}
Jason Wei, Xuezhi Wang, Dale Schuurmans, Maarten Bosma, Fei Xia, Ed~Chi, Quoc~V Le, Denny Zhou, et~al. 2022.
\newblock Chain-of-thought prompting elicits reasoning in large language models.
\newblock In \emph{NeurIPS}.

\bibitem[{Wu et~al.(2023)Wu, Gong, Shou, Liang, and Jiang}]{wu2023large}
Ning Wu, Ming Gong, Linjun Shou, Shining Liang, and Daxin Jiang. 2023.
\newblock Large language models are diverse role-players for summarization evaluation.
\newblock In \emph{NLPCC}.

\bibitem[{Wu et~al.(2020)Wu, Liu, Shi, Sun, Shao, Wu, and Yang}]{wu2020not}
Peng Wu, Jing Liu, Yujia Shi, Yujia Sun, Fangtao Shao, Zhaoyang Wu, and Zhiwei Yang. 2020.
\newblock Not only look, but also listen: Learning multimodal violence detection under weak supervision.
\newblock In \emph{ECCV}.

\bibitem[{Xiao et~al.(2010)Xiao, Hays, Ehinger, Oliva, and Torralba}]{xiao2010sun}
Jianxiong Xiao, James Hays, Krista~A Ehinger, Aude Oliva, and Antonio Torralba. 2010.
\newblock Sun database: Large-scale scene recognition from abbey to zoo.
\newblock In \emph{CVPR}.

\bibitem[{Xu et~al.(2023)Xu, Ren, and Jia}]{xu-etal-2023-contrastive}
Albert Xu, Xiang Ren, and Robin Jia. 2023.
\newblock \href {https://doi.org/10.18653/v1/2023.acl-long.658} {Contrastive novelty-augmented learning: Anticipating outliers with large language models}.
\newblock In \emph{Proceedings of the 61st Annual Meeting of the Association for Computational Linguistics (Volume 1: Long Papers)}, pages 11778--11801, Toronto, Canada. Association for Computational Linguistics.

\bibitem[{Yang et~al.(2024{\natexlab{a}})Yang, Zhou, Li, and Liu}]{yang2024generalized}
Jingkang Yang, Kaiyang Zhou, Yixuan Li, and Ziwei Liu. 2024{\natexlab{a}}.
\newblock \href {https://doi.org/10.1007/s11263-024-02117-4} {Generalized out-of-distribution detection: A survey}.
\newblock \emph{IJCV}.

\bibitem[{Yang et~al.(2024{\natexlab{b}})Yang, Nian, Li, Xu, Li, Li, Xiao, Hu, Rossi, Ding et~al.}]{yang2024ad}
Tiankai Yang, Yi~Nian, Shawn Li, Ruiyao Xu, Yuangang Li, Jiaqi Li, Zhuo Xiao, Xiyang Hu, Ryan Rossi, Kaize Ding, et~al. 2024{\natexlab{b}}.
\newblock Ad-llm: Benchmarking large language models for anomaly detection.
\newblock \emph{arXiv preprint arXiv:2412.11142}.

\bibitem[{Yang et~al.(2024{\natexlab{c}})Yang, Lee, Dariush, Cao, and Lo}]{yang2024follow}
Yuchen Yang, Kwonjoon Lee, Behzad Dariush, Yinzhi Cao, and Shao-Yuan Lo. 2024{\natexlab{c}}.
\newblock Follow the rules: Reasoning for video anomaly detection with large language models.
\newblock \emph{arXiv preprint arXiv:2407.10299}.

\bibitem[{Yin et~al.(2023)Yin, Fu, Zhao, Li, Sun, Xu, and Chen}]{yin2023survey}
Shukang Yin, Chaoyou Fu, Sirui Zhao, Ke~Li, Xing Sun, Tong Xu, and Enhong Chen. 2023.
\newblock A survey on multimodal large language models.
\newblock \emph{arXiv preprint arXiv:2306.13549}.

\bibitem[{Zanella et~al.(2024)Zanella, Menapace, Mancini, Wang, and Ricci}]{zanella2024harnessing}
Luca Zanella, Willi Menapace, Massimiliano Mancini, Yiming Wang, and Elisa Ricci. 2024.
\newblock Harnessing large language models for training-free video anomaly detection.
\newblock In \emph{CVPR}.

\bibitem[{Zhang et~al.(2024{\natexlab{a}})Zhang, Xiao, Liu, Bamler, and Wischik}]{zhang2024your}
Andi Zhang, Tim~Z Xiao, Weiyang Liu, Robert Bamler, and Damon Wischik. 2024{\natexlab{a}}.
\newblock Your finetuned large language model is already a powerful out-of-distribution detector.
\newblock \emph{arXiv preprint arXiv:2404.08679}.

\bibitem[{Zhang and Zuo(2008)}]{pusurvey08isip}
Bangzuo Zhang and Wanli Zuo. 2008.
\newblock Learning from positive and unlabeled examples: A survey.
\newblock In \emph{International Symposiums on Information Processing}.

\bibitem[{Zhang et~al.(2024{\natexlab{b}})Zhang, Xu, Wang, Zuo, Han, Huang, Gao, Wang, and Sang}]{zhang2024holmes}
Huaxin Zhang, Xiaohao Xu, Xiang Wang, Jialong Zuo, Chuchu Han, Xiaonan Huang, Changxin Gao, Yuehuan Wang, and Nong Sang. 2024{\natexlab{b}}.
\newblock Holmes-vad: Towards unbiased and explainable video anomaly detection via multi-modal llm.
\newblock \emph{arXiv preprint arXiv:2406.12235}.

\bibitem[{Zhang et~al.(2023)Zhang, Chen, Xue, Wang, Wang, and Liu}]{zhang2023exploring}
Jiangning Zhang, Xuhai Chen, Zhucun Xue, Yabiao Wang, Chengjie Wang, and Yong Liu. 2023.
\newblock Exploring grounding potential of vqa-oriented gpt-4v for zero-shot anomaly detection.
\newblock \emph{arXiv preprint arXiv:2311.02612}.

\bibitem[{Zhou et~al.(2017)Zhou, Lapedriza, Khosla, Oliva, and Torralba}]{zhou2017places}
Bolei Zhou, Agata Lapedriza, Aditya Khosla, Aude Oliva, and Antonio Torralba. 2017.
\newblock Places: A 10 million image database for scene recognition.
\newblock \emph{TPAMI}.

\bibitem[{Zhou et~al.(2022)Zhou, Yang, Loy, and Liu}]{zhou2022learning}
Kaiyang Zhou, Jingkang Yang, Chen~Change Loy, and Ziwei Liu. 2022.
\newblock Learning to prompt for vision-language models.
\newblock \emph{IJCV}.

\bibitem[{Zhou et~al.(2024)Zhou, Pang, Tian, He, and Chen}]{zhou2024anomalyclip}
Qihang Zhou, Guansong Pang, Yu~Tian, Shibo He, and Jiming Chen. 2024.
\newblock Anomaly{CLIP}: Object-agnostic prompt learning for zero-shot anomaly detection.
\newblock In \emph{ICLR}.

\bibitem[{Zhou(2022)}]{zhou2022rethinking}
Yibo Zhou. 2022.
\newblock Rethinking reconstruction autoencoder-based out-of-distribution detection.
\newblock In \emph{CVPR}.

\bibitem[{Zhu et~al.(2024)Zhu, Cai, Deng, and Wu}]{zhu2024llms}
Jiaqi Zhu, Shaofeng Cai, Fang Deng, and Junran Wu. 2024.
\newblock \href {https://openreview.net/forum?id=JyOGUqYrbV} {Do llms understand visual anomalies? uncovering llm capabilities in zero-shot anomaly detection}.
\newblock In \emph{ACM Multimedia 2024}.

\bibitem[{Zhu and Pang(2024)}]{zhu2024toward}
Jiawen Zhu and Guansong Pang. 2024.
\newblock Toward generalist anomaly detection via in-context residual learning with few-shot sample prompts.
\newblock In \emph{CVPR}.

\bibitem[{Zisselman and Tamar(2020)}]{residualflow20cvpr}
Ev~Zisselman and Aviv Tamar. 2020.
\newblock Deep residual flow for out of distribution detection.
\newblock In \emph{CVPR}.

\bibitem[{Zong et~al.(2018)Zong, Song, Min, Cheng, Lumezanu, Cho, and Chen}]{dagmm18iclr}
Bo~Zong, Qi~Song, Martin~Renqiang Min, Wei Cheng, Cristian Lumezanu, Daeki Cho, and Haifeng Chen. 2018.
\newblock Deep autoencoding gaussian mixture model for unsupervised anomaly detection.
\newblock In \emph{ICLR}.

\bibitem[{Zou et~al.(2022)Zou, Jeong, Pemula, Zhang, and Dabeer}]{zou2022spot}
Yang Zou, Jongheon Jeong, Latha Pemula, Dongqing Zhang, and Onkar Dabeer. 2022.
\newblock Spot-the-difference self-supervised pre-training for anomaly detection and segmentation.
\newblock In \emph{ECCV}.

\end{thebibliography}
\clearpage
\newpage
\appendix

\section{LLM-based Anomaly and OOD Detection: Strengths and Challenges}

\subsection{Strengths}
\noindent\textbf{Zero-shot/Few-shot:} Traditional anomaly and OOD detection methods require extensive training on well-defined normal and in-distribution datasets, which can be both time-consuming and computationally expensive. In contrast, LLMs can perform zero-shot or few-shot reasoning or learning, producing detection results without needing large-scale training \cite{kojima2022large}. This allows researchers to bypass the complex process of data collection and model training, enabling faster anomaly and OOD detection.

\noindent\textbf{Explainability and Interpretability:} 
LLMs possess strong reasoning abilities that can contribute to building explainable systems for anomaly and OOD detection. Traditional methods often rely on scores that offer little insight into the detection process. In contrast, LLMs can provide detailed, interpretable explanations for their detection results, offering valuable insights for future actions. Furthermore, LLMs can be integrated as agents within a system, assisting in planning the next steps when an anomaly or OOD event is detected.

\subsection{Challenges}
\noindent\textbf{Computational Efficiency and Token Limits:} 
A major concern when leveraging LLMs for anomaly and OOD detection is computational inefficiency. Applying LLMs in these tasks often requires complex reasoning which can lead to significant computational overhead. Additionally, many LLMs have input token limits, making it impossible to feed large amounts of data directly into the model. To address this, researchers must carefully design architectures that allow LLMs to process the data effectively. For instance, techniques like Retrieval-Augmented Generation have been used to retrieve the most relevant data to avoid token limit issues \cite{liu2024large}. Moreover, methods such as model pruning and knowledge distillation should be considered to reduce computational costs while maintaining high accuracy. 

\noindent\textbf{Domain Knowledge:} 
While LLMs are trained on vast and diverse datasets, they may lack specific domain expertise needed for certain anomaly and OOD detection tasks. To enhance performance in these specialized domains, incorporating domain knowledge into the LLMs is crucial. One strategy is injecting domain knowledge into prompts to guide the LLM's understanding. Another approach involves using adapters or fine-tuning to better tailor the LLMs to domain-specific problems, ensuring they perform well in specialized tasks.

\noindent\textbf{Hallucination and Trustworthiness:}
LLMs can sometimes produce inaccurate or fabricated information, a phenomenon known as hallucination. In the context of anomaly and OOD detection, hallucinations pose a significant risk, potentially leading to incorrect or misleading results. To mitigate this, researchers need to work on reducing hallucination rates and improving the trustworthiness of the model. Manual checks may still be necessary in critical applications, as LLMs should be seen as assistants rather than sole decision-makers.

\section{Anomaly Detection Research Roadmap}
Anomaly detection has evolved significantly from traditional statistical methods to deep learning approaches and more recently to Large Language Model (LLM)-based methods. These advancements have expanded the range of applications across diverse data modalities and downstream tasks. Following \citet{yang2024generalized}, traditional methodologies can be grouped into density-based, reconstruction-based, distance-based, and classification-based methods. Below is an overview of these traditional approaches, followed by a discussion of the advantages and challenges of LLM-based anomaly detection.

\subsection{Traditional Methods}

\noindent\textbf{Density-based Methods.}
Density-based methods model the distribution of normal data and detect anomalies by evaluating how well a sample fits this modeled distribution. The underlying assumption is that normal data is more likely to have a higher likelihood under the distribution, while anomalous data will have a lower likelihood.

Parametric density estimation assumes a predefined form for the distribution, such as a multivariate Gaussian or Poisson distribution~\cite{anomalyparametric98pami,mahalanobis18jesp,poisson16isi}. These methods perform well when the data distribution adheres to the parametric assumption but can struggle with more complex cases. Non-parametric density estimation methods, such as histograms and kernel density estimation (KDE), offer more flexibility by not assuming a fixed parametric form, making them suitable for handling more complex distributions \cite{kde62math,anomalykde18tkde}.
Modern deep learning techniques enhance density estimation by learning high-quality feature representations. Methods such as autoencoders (AE), variational autoencoders (VAE), and flow-based models are commonly used \cite{ae91aiche,vae13arxiv,gan14nips}. 

\noindent\textbf{Reconstruction-based Methods.}
Reconstruc-\\tion-based methods operate on the assumption that models trained on normal data will reconstruct those samples accurately, while anomalous data will result in higher reconstruction errors. This discrepancy in reconstruction performance is used to identify anomalies. Common approaches include sparse reconstruction, where normal samples are represented by a small set of basis functions, while anomalies are not. Autoencoders and variational autoencoders are often employed to capture these differences in reconstruction errors  \cite{ae91aiche,vae13arxiv}.

Recent advancements have sought to reduce the computational costs associated with reconstruction-based methods, for example by focusing on reconstructing hidden features or masking parts of the input \cite{gpnd18nips}. These improvements enhance both the accuracy and efficiency of anomaly detection, without requiring pixel-level reconstruction.

\noindent\textbf{Distance-based Methods.}
Distance-based methods detect anomalies by measuring the distance between test samples and reference points, such as class prototypes or centroids. Anomalies are expected to be further from these reference points than normal samples \cite{tian2014anomaly}.

\noindent\textbf{Classification-based Methods.}
Classification-based methods treat anomaly detection as a supervised learning problem. In one-class classification (OCC)~\cite{occ02tax}, the goal is to learn a decision boundary that encompasses the normal data, with any data points outside this boundary being classified as anomalies.

DeepSVDD \cite{deepsvdd18icml} is a notable deep learning method for OCC, where a deep network learns a compact representation of the normal class. Semi-supervised approaches, such as positive-unlabeled (PU) learning~\cite{pusurvey08isip,pusurvey20ml,pusurvey19iisa}, are also used when only a subset of the normal data is labeled, with the rest being unlabeled. Self-supervised learning methods have been proposed as well, using pretext tasks such as contrastive learning or future frame prediction to identify anomalies~\cite{csi20nips}.

\subsection{LLM-based Anomaly Detection}
The advent of Large Language Models (LLMs) like GPT, as well as multimodal LLMs, has introduced new possibilities for anomaly detection. LLMs offer zero-shot and few-shot learning capabilities, allowing anomalies to be detected with minimal or no task-specific training. This is particularly valuable in scenarios where labeled data is scarce or unavailable. LLM-based approaches benefit from their pre-trained knowledge and can adapt to various data modalities, including images, and videos. By using natural language processing capabilities, LLMs can provide explainability in anomaly detection, offering reasoning as to why a particular instance is flagged as anomalous.

However, LLM-based methods come with certain challenges, including high computational costs and limitations related to token size. These models are computationally intensive to run and may struggle with long input sequences, which necessitates techniques such as retrieval-augmented generation (RAG) or model pruning to manage these constraints.

\subsection{Comparison: Traditional vs. LLM-based Anomaly Detection}
When comparing traditional and LLM-based anomaly detection methods, several key differences emerge:

\begin{itemize}[leftmargin=*,itemsep=0.2pt,topsep=1.pt]
    \item \textit{Assumptions}: Traditional methods often rely on predefined assumptions about the data distribution, such as the parametric forms used in density-based methods. In contrast, LLM-based approaches are better equipped to generalize across a variety of tasks.
  
    \item \textit{Data Requirements}: Traditional methods, particularly those based on deep learning, usually require large labeled datasets for training. LLM-based methods excel in zero-shot or few-shot settings, enabling them to detect anomalies with minimal task-specific data.
  
    \item \textit{Explainability}: Traditional methods lack the ability to explain their decisions in natural language. LLM-based approaches can not only detect anomalies but also provide natural language explanations, which improves transparency and trustworthiness for further steps.
  
    \item \textit{Computational Efficiency}: Traditional methods are generally more computationally efficient compared to LLM-based methods, especially when fine-tuning models. However, LLM-based approaches offer greater flexibility and can handle a wider range of tasks, though at the cost of higher computational resources. Moreover, LLMs do not require extensive data preparation and training from scratch, which can offset the computational overhead in certain scenarios.
  
    \item \textit{Generalization}: LLM-based methods are highly adaptable, capable of processing different types of data, such as text and images. In contrast, traditional methods often need custom architectures tailored to the specific data modality.

\end{itemize}

In conclusion, while traditional anomaly detection methods remain effective and computationally efficient, LLM-based methods provide greater flexibility, generalization, and explainability. This makes LLM-based approaches increasingly valuable in modern, complex anomaly detection tasks.

\section{Out-of-Distribution Detection Research Roadmap}

Compared to Anomaly Detection (AD), Out-of-distribution (OOD) detection emerged in 2017 and has since received increasing attention. OOD detection is critical for ensuring the reliability and safety of machine learning models by identifying samples that fall outside the distribution of the training data. Following \citet{yang2024generalized}, traditional OOD detection methods can be categorized into classification-based, density-based, distance-based, and reconstruction-based methods. These approaches vary in how they define and detect OOD samples, with each showing strengths depending on data characteristics and the task at hand.

\subsection{Traditional Methods}

\noindent\textbf{Classification-based Methods.}
Classification-based OOD detection methods rely on the outputs of neural networks , typically using the softmax probabilities of a classifier to determine whether a sample is in-distribution (ID) or OOD. The most common baseline is the Maximum Softmax Probability (MSP) method, which flags samples with lower softmax scores as OOD~\cite{msp17iclr}. This has led to more advanced techniques that either post-process the classification outputs or modify the training process to improve OOD detection performance~\cite{liu2020energy}. Given its alignment with classification tasks, this approach remains one of the most prominent methods for OOD detection.

\noindent\textbf{Density-based Methods.}
Density-based methods explicitly model the distribution of in-distribution (ID) data using probabilistic models, assuming that OOD samples will lie in low-density regions~\cite{dagmm18iclr,autogres19cvpr,gpnd18nips,adgan18ecml,advrec18cvpr}. Techniques such as class-conditional Gaussian models allow for probabilistic modeling of ID classes, while flow-based models are also used for density estimation~\cite{flowreview20pami,residualflow20cvpr,glow18nips,pixelcnn16icml}. However, density-based methods sometimes assign higher likelihoods to OOD samples, leading to challenges in reliability. Methods such as likelihood ratio-based approaches and ensembles have been proposed to address these issues, though these approaches tend to be computationally expensive and often fall behind classification-based methods in performance.

\noindent\textbf{Distance-based Methods.}
Distance-based methods operate under the assumption that OOD samples are farther from the centroids or prototypes of in-distribution classes in feature space. A popular parametric approach is to use Mahalanobis distance to compute the distance between test samples and class centroids~\cite{lee2018simple}, whereas non-parametric methods are increasingly favored for their flexibility and simplicity~\cite{sun2022knn}. These methods use various distance metrics—such as Euclidean distance and geodesic distance—to detect OOD samples. 

\noindent\textbf{Reconstruction-based Methods.}
Reconstruc-\\tion-based methods leverage encoder-decoder models to reconstruct input samples and detect OOD samples by measuring reconstruction error. The premise is that models trained on ID data will exhibit lower reconstruction errors for ID samples and higher errors for OOD samples. \cite{zhou2022rethinking}. 

\subsection{LLM-based OOD Detection}
Large Language Models (LLMs) and Multi-Modal LLMs (MLLMs) have transformed Out-of-Distribution (OOD) detection by leveraging pre-trained models like CLIP to perform downstream detection tasks. These models are capable of detecting OOD samples in zero-shot or few-shot settings, meaning they can generalize to unseen data with little to no additional training. This represents a shift from traditional OOD detection methods, which typically rely on training classifiers using the entire in-distribution (ID) dataset.

Incorporating the internal knowledge of pre-trained MLLMs, the field is progressing towards even greater computational efficiency, where minimal or no training data is needed. This ability to operate with limited data while maintaining performance makes LLM-based OOD detection especially appealing for real-world applications.

\subsection{Comparison of Traditional vs. LLM-based OOD Detection}
The shift from traditional OOD detection methods to LLM-based approaches marks a fundamental change in how OOD detection is defined and executed. Traditional methods, such as classification-based approaches like Maximum Softmax Probability (MSP) or distance-based techniques like Mahalanobis distance, rely heavily on task-specific training and typically require large amounts of in-distribution (ID) data. These methods often define OOD detection in the context of post-processing or retraining models on ID data to differentiate between in- and out-of-distribution samples.

In contrast, LLM-based methods redefine OOD detection by leveraging pre-trained models, allowing them to detect OOD samples without extensive task-specific training. This results in a significant shift in the OOD detection paradigm, moving towards zero-shot and few-shot learning, where models can generalize to new tasks with minimal or no additional training. Key differences include:
\begin{itemize}[leftmargin=*,itemsep=0.2pt,topsep=1.pt]
    \item \textit{Performance}: LLM-based methods often outperform traditional methods in zero-shot and few-shot scenarios, where limited labeled data is available. Traditional methods struggle without substantial in-distribution data and retraining.

    \item \textit{Flexibility}: LLM-based approaches are highly adaptable to new tasks and datasets due to their reliance on vast pre-trained knowledge, while traditional methods require significant retraining for new domains or data types.
    
    \item \textit{Efficiency}: Traditional OOD methods rely on explicitly training models with in-distribution data, while LLM-based methods redefine the problem by leveraging pre-existing knowledge in zero-shot or few-shot settings, minimizing the need for retraining or task-specific data preparation.
\end{itemize}

In conclusion, the transition from traditional OOD detection methods to LLM-based approaches represents a shift from task-specific training and rigid models to more flexible, generalizable systems that can handle a wider variety of tasks with minimal additional training. As research continues, a hybrid approach combining the efficiency of traditional methods with the flexibility LLM-based models may offer the most robust solution for diverse OOD detection challenges.

\section{Quantitative Analysis and Comparison}
While our primary goal is to conduct a systematic literature review of existing methods for anomaly and out-of-distribution (OOD) detection tasks, we acknowledge that including quantitative analysis and comparisons is valuable for understanding the practical implications of these methods. 
\begin{table}[t]
    \centering
    \scalebox{0.75}{
    \begin{tabular}{c c c }
        \hline
        \hline
        \textbf{Method} & \textbf{MVTec AD} & \textbf{VisA} \\ 
        &  AUC $\uparrow$  &  AUC $\uparrow$  \\ 
        \hline
        \textit{Zero-shot} & ~ & ~   \\
        \textbf{CoOp}~\cite{zhou2022learning}$^\dag$ & 88.8 & 62.8 \\
        \textbf{CLIP-AC}~\cite{radford2021learning}$^\dag$ & 71.5 & 65.0 \\
        \textbf{VAND}~\cite{chen2023april}$^\dag$ & 86.1 & 78.0 \\
        \textbf{AnomalyCLIP}~\cite{zhou2024anomalyclip}$^\dag$ & \textbf{91.5} & 82.1 \\
        \textbf{CLIP-AD}~\cite{liznerski2022exposing}$^\dag$ & 90.9 & 79.2 \\
        \textbf{FiLo}~\cite{gu2024filo}$^\dag$ & 91.2 & \textbf{83.9} \\
        \hline
        \hline
        \textit{One-shot} & ~ & ~   \\
        \textbf{SPADE}~\cite{park2019semantic} & 81.0 & 79.5 \\
        \textbf{PaDiM}~\cite{defard2021padim} & 76.6 & 62.8 \\
        \textbf{PatchCore}~\cite{roth2022towards} & 83.4 & 79.9 \\
        \textbf{WinCLIP}~\cite{jeong2023winclip}$^\dag$ & 93.1 & 83.8 \\
        \textbf{AnomalyGPT}~\cite{gu2024anomalygpt}$^\dag$ & 94.1 & 87.4 \\
        \textbf{AnomalyDINO-S}~\cite{damm2024anomalydino}$^\dag$ & \textbf{96.6} & \textbf{87.4} \\
        \hline
    \end{tabular}
    }
    \caption{Anomaly detection results for MVTec AD and VisA (image-level). Bold indicates the best performance. The methods marked with $\dag$ are using MLLMs as backbones. The results are are sourced from \cite{zhou2024anomalyclip,gu2024anomalygpt}}
    \label{tab:performance_comparison_combined}
\end{table}

\begin{table*}[t]
    \centering
    \resizebox{\textwidth}{!}{
    \begin{tabular}{c cc  cc   cc  cc |cc}
        \hline
        \hline
        \multirow{2}{*}{\textbf{Method}} & \multicolumn{2}{c}{\textbf{Texture}} & \multicolumn{2}{c}{\textbf{iNaturalist}} & \multicolumn{2}{c}{\textbf{Places}} & \multicolumn{2}{c|}{\textbf{SUN}} & \multicolumn{2}{c}{\textbf{Avg}} \\
        ~  &  AUC $\uparrow$  & FPR95 $\downarrow$ &  AUC $\uparrow$  & FPR95 $\downarrow$ &  AUC $\uparrow$  & FPR95 $\downarrow$ &  AUC $\uparrow$  & FPR95 $\downarrow$ &  AUC $\uparrow$  & FPR95 $\downarrow$ \\
        \specialrule{0em}{1pt}{1pt}
        \hline
        \specialrule{0em}{1pt}{1pt}
        \textit{Traditional posthoc methods} & ~ & ~ & ~ & ~ & ~ & ~ & ~ & ~ & ~ & ~ \\
        \textbf{MSP}~\cite{hendrycks2019scaling}$^\dag$ & 74.84 & 73.66 & 77.74 & 74.57 & 72.18 & 79.12 & 73.97 & 76.95 & 74.98 & 76.22 \\
        \textbf{MaxLogit}~\cite{hendrycks2019scaling}$^\dag$ & 88.63 & 48.72 & 88.03 & 60.88 & 87.45 & 55.54 & 91.16 & 44.83 & 88.82 & 52.49 \\
        \textbf{Energy}~\cite{liu2020energy}$^\dag$ & 88.22 & 50.39 & 87.18 & 64.98 & 87.33 & 57.40 & 91.17 & 46.42 & 88.48 & 54.80 \\
        \textbf{ReAct}~\cite{sun2021react}$^\dag$ & 88.13 & 49.88 & 86.87 & 65.57 & 87.42 & 56.85 & 91.04 & 46.17 & 88.37 & 54.62 \\
        \textbf{ODIN}~\cite{liang2018enhancing} $^\dag$ & 87.85 & 51.67 & 94.65 & 30.22 & 85.54 & 55.06 & 87.17 & 54.04 & 88.80 & 47.75 \\
        \hline
        \hline
        \textit{Without Tuning methods} & ~ & ~ & ~ & ~ & ~ & ~ & ~ & ~ & ~ & ~ \\
        \textbf{MCM}~\cite{ming2022delving}$^\dag$ & 86.11 & 57.77 & 94.61 & 30.91 & 89.77 & 44.69 & 92.57 & 34.59 & 90.76 & 42.74 \\
        \textbf{NegLabel} \cite{jiang2024negative} & 90.22 & 43.5 & \textbf{99.49} & \textbf{1.91}&  91.64 & 35.59 & 95.49 & \textbf{20.53} & 94.21 & 25.40 \\
        \textbf{EOE} ~\cite{cao2024envisioning} & 57.53 &85.64 & 97.52 & 12.29& 95.73 & 20.40 & 92.95 & 30.16 & 92.96 & 30.09 \\
        \hline
        \textit{With Tuning methods} & ~ & ~ & ~ & ~ & ~ & ~ & ~ & ~ & ~ & ~ \\
        \textbf{CoOp}~\cite{zhou2022learning} & 89.47 & 45.00 & 93.77 & 29.81 & 90.58 & 40.11 & 93.29 & 40.83 & 91.78 & 51.68 \\
        \textbf{LoCoOp}~\cite{miyai2024locoop}$^\dag$ & 90.19 & 42.28 & 96.86 & 16.05 & 91.98 & 32.87 & 95.07 & 23.44 & 93.52 & 28.66 \\
        \textbf{CLIPN}~\cite{wang2023clipn}$^\dag$ & 90.93 & 40.83 & 95.27 & 23.94 & 92.28 & 33.45 & 93.92 & 26.17 & 93.10 & 31.10 \\
        \textbf{NegPrompt} ~\cite{nie2024outofdistribution} & \textbf{91.60} & \textbf{35.21} & 98.73 & 6.32 & \textbf{93.34} & \textbf{27.60} & \textbf{95.55} & 22.89 & \textbf{94.81} & \textbf{23.01} \\
        \hline
    \end{tabular}
    }
    \caption{Comprehensive OOD detection results for ImageNet-1K as ID dataset. The black bold indicates the best performance. The results marked with $\dag$ are sourced from \cite{wang2023clipn} and \cite{miyai2024locoop}. Others are sourced from their original papers \cite{wang2023clipn,jiang2024negative,cao2024envisioning,nie2024outofdistribution}}
    \label{tab:comprehensive_results_with_EOE}
\end{table*}
\subsection{Anomaly Detection}
Table \ref{tab:performance_comparison_combined}
 presents the quantitative results for image-level anomaly detection on two widely used benchmarks, MVTec AD \cite{8954181} and VisA \cite{zou2022spot}. The results are sourced from the original papers. We ensured that the experiments across different methods used the same dataset and learning settings for a fair comparison.

\subsection{OOD Detection}
Table \ref{tab:comprehensive_results_with_EOE} provides an overview of out-of-distribution (OOD) detection results using ImageNet-1K as the in-distribution (ID) dataset. The methods are evaluated across multiple OOD datasets, including Texture, iNaturalist, Places, and SUN~\cite{van2018inaturalist, zhou2017places, xiao2010sun, texture}, with performance measured using AUC (Area Under the ROC Curve) and FPR95 (False Positive Rate at 95\% True Positive Rate).

\section{Evaluation Metrics}
\label{sec:metrics}
The primary goal of both anomaly detection and out-of-distribution (OOD) detection is to differentiate between normal/in-distribution (ID) samples and abnormal/out-of-distribution (OOD) samples, framing the problem as a binary classification task. Several common metrics are used to evaluate the performance of detectors:

\noindent\textbf{AUROC (Area Under the Receiver Operating Characteristic curve):} This metric evaluates a detector's overall ability to distinguish between ID or normal and OOD or anomalous samples. The ROC curve plots the true positive rate (TPR) against the false positive rate (FPR), where:
\[
TPR = \frac{TP}{TP + FN}, \quad FPR = \frac{FP}{FP + TN}
\]
Here, TP (true positives), TN (true negatives), FP (false positives), and FN (false negatives) correspond to the detector’s correct and incorrect classifications. 

\noindent\textbf{AUPR (Area Under the Precision-Recall curve):} The AUPR metric is particularly useful for cases where there is class imbalance, as AUROC can be biased in such situations. The Precision-Recall curve plots precision against recall, where:
\[
Precision = \frac{TP}{TP + FP}, \quad Recall = \frac{TP}{TP + FN}
\]

\noindent\textbf{FPR@N (False Positive Rate at TPR = N\%):} This metric evaluates the probability of misclassifying an OOD or anomalous sample as ID or normal when the true positive rate (TPR) is set at a specified value, commonly 90\% or 95\%. This is crucial for real-world deployments where achieving high accuracy on ID samples is important, while also minimizing false positives for OOD or anomaly detection.

\noindent\textbf{F1 Score:} The F1 score is a harmonic mean of precision and recall, providing a balanced evaluation of a model's performance across both metrics. It is particularly useful in scenarios where there is an imbalance between the positive and negative classes, as it gives a single metric that reflects both false positives and false negatives.

The F1 score is calculated as:
\[
F1 = 2 \times \frac{Precision \times Recall}{Precision + Recall}
\]
where:
\[
Precision = \frac{TP}{TP + FP}, \quad Recall = \frac{TP}{TP + FN}
\]
Precision measures the proportion of true positive predictions out of all positive predictions made by the model, while recall measures the proportion of actual positives correctly identified by the model. There are two main variations of the F1 score:
\begin{itemize}[leftmargin=*,itemsep=0.2pt,topsep=1.pt]
    \item \textbf{Macro F1 Score}: This version computes the F1 score independently for each class (ID and OOD, or normal and anomalous) and then takes the average across classes. It treats all classes equally, making it particularly useful when the class distribution is imbalanced.
    \item \textbf{Micro F1 Score}: This version aggregates the contributions of all classes to calculate the F1 score, considering the total number of true positives, false positives, and false negatives across all classes. It is more sensitive to the performance on the larger class.
\end{itemize}

A high F1 score indicates that the model maintains a good balance between precision (minimizing false positives) and recall (minimizing false negatives), which is critical in practical OOD detection and anomaly detection tasks.

\section{General Guidelines}
We provide general guidelines for selecting appropriate approaches for anomaly or out-of-distribution (OOD) detection, considering key factors such as data modality, efficiency, explanation, and optimization.

\paragraph{Data Modality:} The choice of approach is strongly influenced by the type of data being analyzed. For textual data, prompting-based methods may not always offer meaningful interpretations of anomalies or OOD detection, particularly when trying to understand patterns in semantic spaces. In these cases, generating embeddings from LLMs and applying specialized post-hoc detection techniques can lead to better results. Fine-tuning LLMs to produce more relevant embeddings may further enhance detection accuracy. In the case of numerical data, such as time series or tabular data, prompting-based methods have been explored, though they often require carefully designed prompts or fine-tuning to capture the underlying structure of the data. For vision data, including images and videos, the development of multimodal LLMs offers greater flexibility. Both prompting-based and contrasting-based methods can be highly effective, as they are capable of handling the diverse characteristics of multimodal data.

\paragraph{Efficiency:} Efficiency is a crucial consideration when choosing between prompting-based and contrasting-based methods. Prompting-based methods can be inefficient for high-precision numerical tasks, as the conversion of numerical values into text results in excessively long input sequences. This inefficiency can become a bottleneck for tasks involving long-term predictions or large datasets, where the computational overhead of generating long outputs becomes significant. In contrast, contrasting-based methods are more efficient for detection tasks for image. By utilizing contrastive objectives to distinguish between positive and negative samples, these methods excel at zero-shot classification and are computationally more efficient, especially for handling multimodal anomaly and OOD detection. Researchers can also explore the use of contrasting-based approaches to numerical data modalities with their visual representations. 

\paragraph{Explanation:} In fields where transparency and interpretability are critical, explanation plays a pivotal role in model selection. LLMs offer a unique advantage by not only detecting anomalies but also generating human-like explanations. This capability is especially valuable in domains where actionable insights and interpretable results are essential for decision-making. Research should focus on enhancing the ability of LLMs to explain their outputs in a way that aligns with domain-specific requirements for clarity and transparency.

\paragraph{Optimization:} The optimization strategy is highly dependent on the specific data modality and the nature of the detection task. For prompting-based methods, parameter-efficient tuning techniques—such as Low-Rank Adaptation are essential for improving model performance without incurring high computational costs. These techniques enable models to adapt to new tasks efficiently while maintaining their generalization capabilities. This allows the broad generalization capabilities of LLMs to be leveraged while optimizing for specific domain-related tasks, leading to improved accuracy and reduced computational overhead.

\end{document}